\documentclass[10pt,twocolumn,letterpaper]{article}

\usepackage[pagenumbers]{cvpr} 
\usepackage{tikz}
\usepackage{arydshln}
\usepackage{multirow}
\usepackage{float}
\definecolor{cvprblue}{rgb}{0.21,0.49,0.74}
\usepackage[pagebackref,breaklinks,colorlinks,allcolors=cvprblue]{hyperref}

\def\paperID{16302} 
\def\confName{CVPR}
\def\confYear{2025}

\title{MeasureNet: Measurement Based Celiac Disease Identification}

\author{
Aayush Kumar Tyagi$^{1}$, Vaibhav Mishra$^{1}$, Ashok Tiwari$^{3}$, Lalita Mehra$^{3}$ ,\\ Prasenjit Das$^{3}$,
Govind Makharia$^{3}$, Prathosh AP$^{2}$, Mausam$^{1}$\\
$^{1}$IIT Delhi ~~~~~~$^{2}$IISc, Bangalore
~~~~~~$^{3}$AIIMS, New Delhi\\
{\small \{tyagiaayushkumar, vaibhavxxmishra, ashoka.pareek, prasenaiims, govindmakharia, prathoshap\}@gmail.com, mausam@cse.iitd.ac.in
}
}

\begin{document}

\newcommand{\model}{\textsc{MeasureNet}}
\maketitle

\begin{abstract}

Celiac disease is an autoimmune disorder triggered by the consumption of gluten. It causes damage to the villi, the finger-like projections in the small intestine that are responsible for nutrient absorption. Additionally, the crypts, which form the base of the villi, are also affected, impairing the regenerative process. The deterioration in villi length, computed as the villi-to-crypt length ratio, indicates the severity of celiac disease. However, manual measurement of villi-crypt length can be both time-consuming and susceptible to inter-observer variability, leading to inconsistencies in diagnosis. While some methods can perform measurement as a post-hoc process, they are prone to errors in the initial stages. This gap underscores the need for pathologically driven solutions that enhance measurement accuracy and reduce human error in celiac disease assessments.

Our proposed method, MeasureNet, is a pathologically driven polyline detection framework incorporating  polyline localization and object-driven losses specifically designed for measurement tasks. Furthermore, we leverage segmentation model to provide auxiliary guidance about crypt location when crypt are partially visible. To ensure that model is not overdependent on segmentation mask we enhance model robustness through a mask feature mixup technique. Additionally, we introduce a novel dataset for grading celiac disease, consisting of 750 annotated duodenum biopsy images. MeasureNet achieves an 82.66\% classification accuracy for binary classification and 81\% accuracy for multi-class grading of celiac disease. Code: \url{https://github.com/dair-iitd/MeasureNet}


\end{abstract}
\section{Introduction}


\begin{figure}[t]
\centering

\begin{tikzpicture}[scale=1.0,transform shape, picture format/.style={inner sep=0.5pt}]
  \node[picture format]                   (A1)   at (0,0)            {\includegraphics[width=1.05in, height = 1.1in]{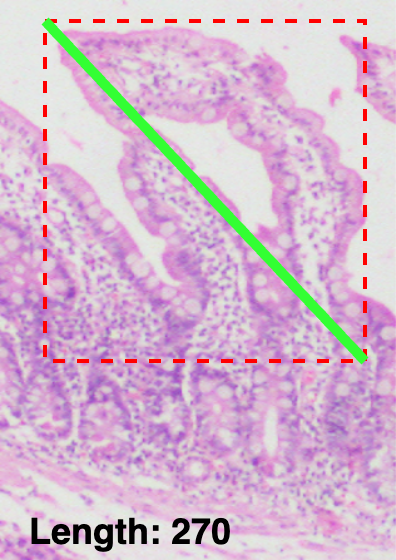}};
  \node[picture format,anchor=north west] (A2)   at (A1.north east)       {\includegraphics[width=1.05in, height = 1.1in]{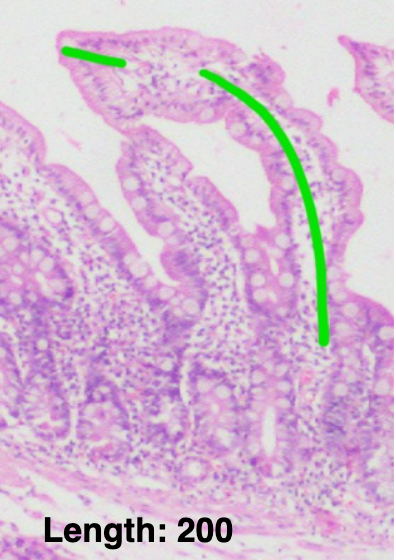}};
  \node[picture format,anchor=north west] (A3) at (A2.north east) {\includegraphics[width=1.05in, height = 1.1in]{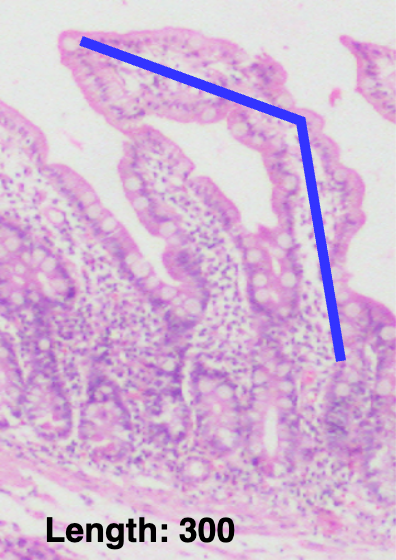}};
  \node[anchor=south] (N1) at (A1.north) {\bfseries Bounding Box};
  \node[anchor=south] (N2) at (A2.north) {\bfseries Segmentation};
  \node[anchor=south] (N3) at (A3.north) {\bfseries Polyline};

\end{tikzpicture}

\caption{Challenges in length measurement: Measuring length along the diagonal or major axis of bounding box can overlook the curvature of the villi, leading to inaccurate assessments. Segmentation masks can produce discontinuous predictions, resulting in errors in length computation. In contrast, polyline-based measurement provides a more accurate length estimate by adapting to the curvature of the villi, effectively capturing true structure.}
\label{fig:measurement_error}
\end{figure}
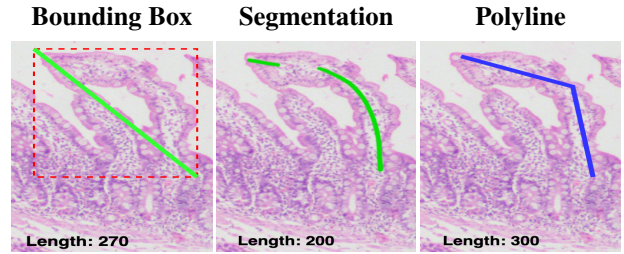


Celiac Disease (CeD) is an autoimmune disorder triggered by consuming gluten, a protein found in wheat, barley, and rye, affecting approximately 1 in 100 people worldwide \cite{vohra2016celiac}. Duodenal biopsy is a primary diagnostic tool for identifying histological changes in CeD patients. Villous atrophy, characterized by the deterioration of nutrient-absorbing villi, can lead to serious complications such as malnutrition, which can be fatal in severe cases. Pathologists assess CeD using the Q-histology score as the primary indicator \cite{tyagi2023degpr} and further evaluate disease severity by calculating the villi-to-crypt length ratio \cite{das2019quantitative}. However, this measurement process is labor-intensive and prone to significant inter- and intra-observer variability \cite{corazza2007comparison, ensari2010gluten}. Additionally, length measurement has applications beyond CeD, including assessing tumor size via the length-to-width ratio \cite{taniyama2021evaluating} and evaluating lower limb alignment \cite{moon2023deep}. These needs highlights the need of a model that ensures accurate and reliable measurement.

\begin{figure}[t]
\centering

\begin{tikzpicture}[scale=1.0,transform shape, picture format/.style={inner sep=0.5pt}]

  \node[picture format]                   (A1)   at (0,0)            {\includegraphics[width=0.75in, height = 0.75in]{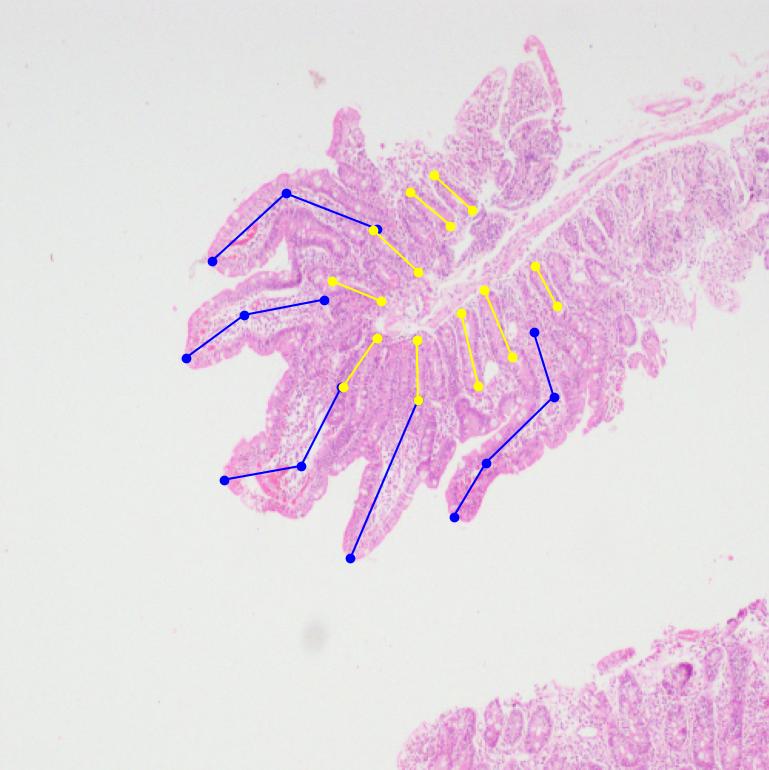}};
  \node[picture format,anchor=north]      (C1) at (A1.south) {\includegraphics[width=0.75in, height = 0.75in]{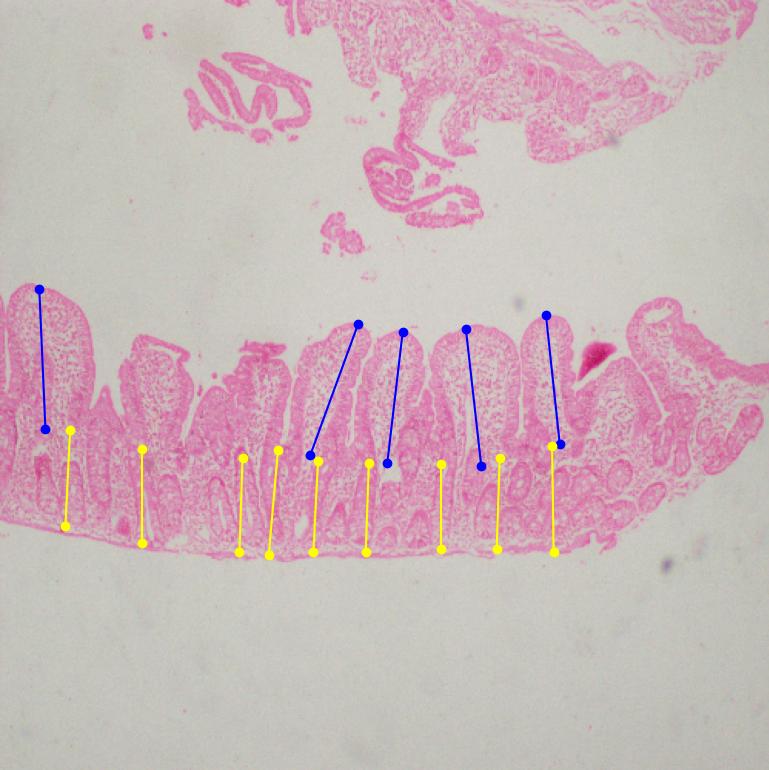}};

  \node[picture format,anchor=north west] (A2)   at (A1.north east)       {\includegraphics[width=0.75in, height = 0.75in]{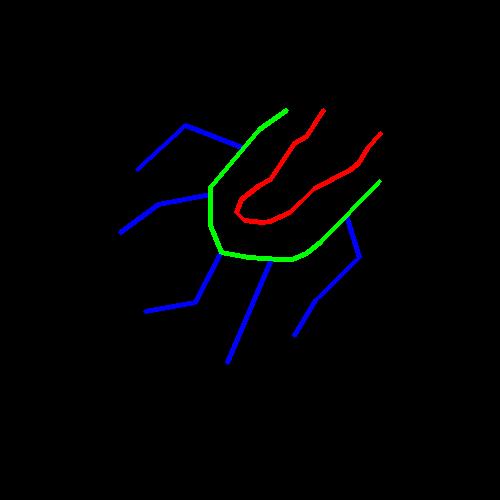}};
  \node[picture format,anchor=north]      (C2) at (A2.south) {\includegraphics[width=0.75in, height = 0.75in]{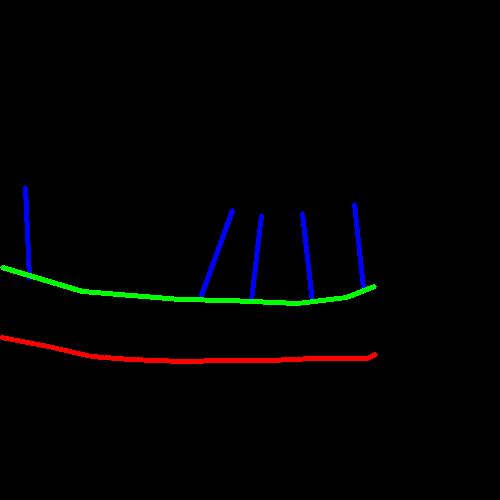}};

  \node[picture format,anchor=north west] (A3) at (A2.north east) {\includegraphics[width=0.75in, height = 0.75in]{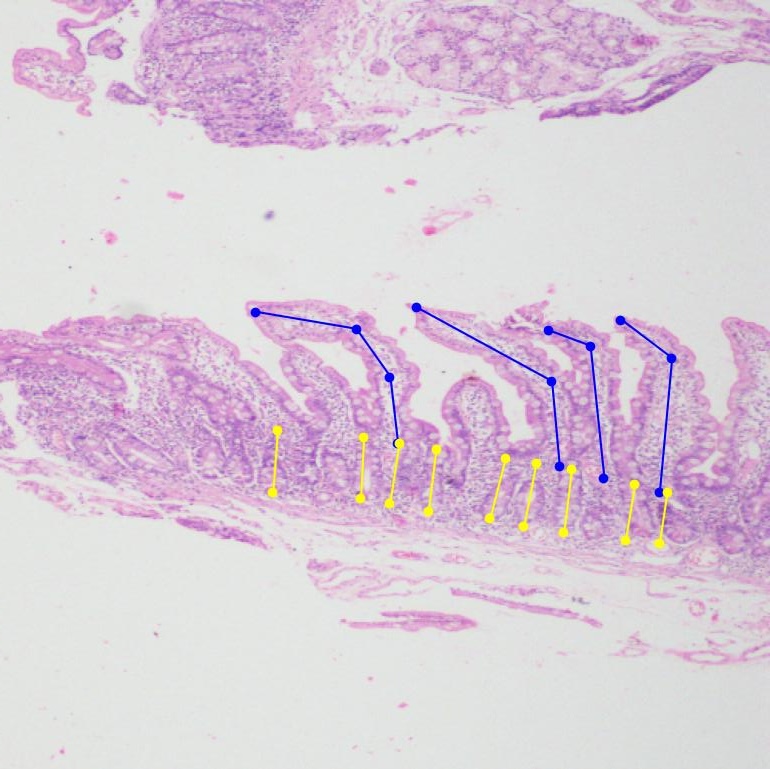}};
   \node[picture format,anchor=north]      (C3) at (A3.south)      {\includegraphics[width=0.75in, height = 0.75in]{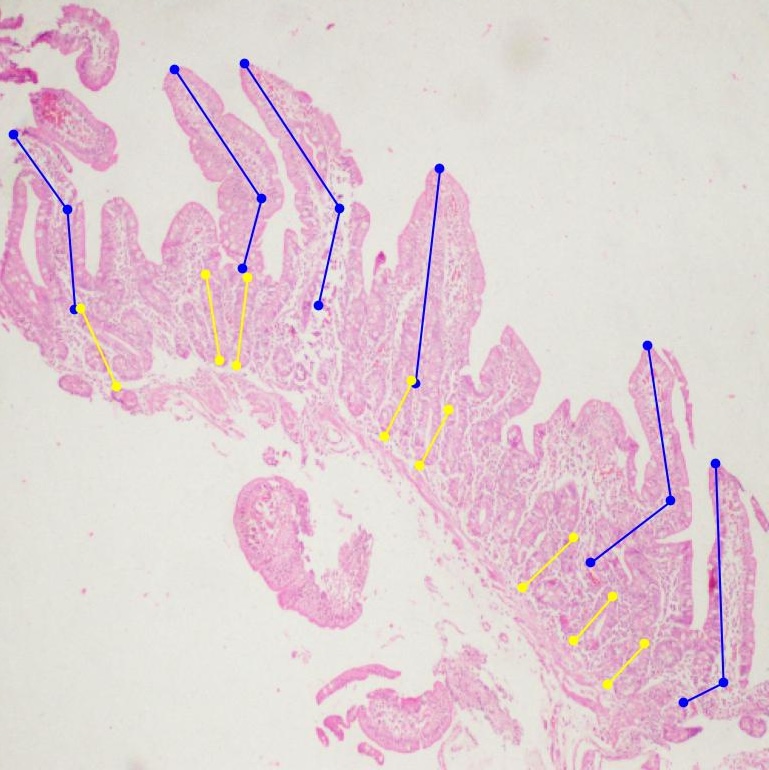}};

  \node[picture format,anchor=north west] (A4) at (A3.north east) {\includegraphics[width=0.75in, height = 0.75in]{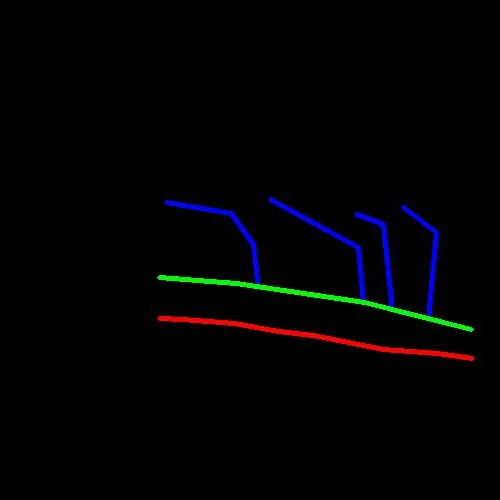}};
  \node[picture format,anchor=north]      (C4) at (A4.south)      {\includegraphics[width=0.75in, height = 0.75in]{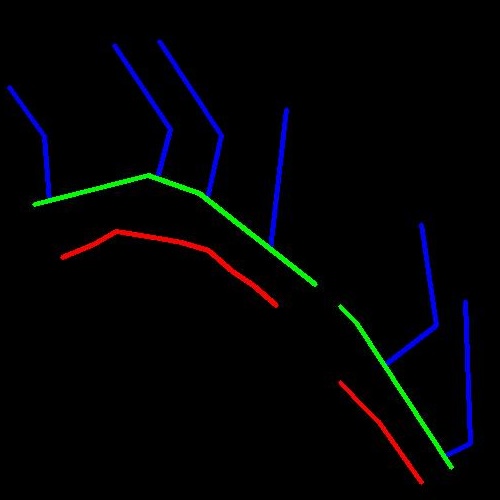}};

\end{tikzpicture}
\caption{Duodenal biopsy of celiac patients, indicating Villi (blue) and Crypt (yellow). Since crypts are not always fully visible, an approximate crypt depth can be estimated as distance between villi shoulder (green) and crypt border (red).}
\label{fig:dataset_images}
\end{figure}

To compute the villi-to-crypt length ratio, precise length measurement is essential. The measurement problem can be framed as a segmentation task \cite{moon2023deep}, where images are processed to generate segmentation maps of object (villi-crypt). The measurement process then becomes a post-hoc task, involving contour identification followed by quantification. However, segmentation models \cite{xie2021segformer, bao2021beit} struggle with generating consistent segmentation masks, leading to discontinuous predictions resulting in post-processing errors. Detection-based approaches, such as lane or line segment detection models designed for natural images, also perform poorly on duodenal biopsy images, where villi-crypt vary in orientation and boundaries are often indistinct (see Fig \ref{fig:dataset_images}). Due to the non-trivial nature of the problem, accuracy in measurement can be improved by integrating pathological insights—such as recognizing that crypts lie between the villi shoulder and the crypt border, as shown in Fig \ref{fig:dataset_images}. However, a key challenge remains in embedding these expert insights within detection model.

We represent each villi or crypt as a polyline — a line segment composed of multiple points—and frame the task as a polyline detection problem. We propose \model{}, a polyline detection framework which is build on DINO-DEtection TRansformer (DINO-DETR) \cite{zhang2022dino}. \model{} leverage general purpose transformer architecture and enhance the performance for detecting fine-grained geometric structures like polylines. Specifically, to CeD, where crypts are not clearly visible, we make approximate inference from the global attributes like villi shoulder and crypt border (see Fig \ref{fig:dataset_images}). For doing so, we use segmentation model like Segformer \cite{xie2021segformer}. We provide the predicted segmentation mask from segmentation model as auxiliary information for detection model to learn from it. Further, to ensure that detection model is do not suffer from exposure bias \cite{arora2022exposure}, i.e. not being exposed to mistakes during training time, we augment the segmentation mask features using mixup between strong and weak segmentation mask.

In addition to our method, we contribute a novel dataset, Celiac disease Detection and grading through measurement (CeDeM). CeDeM consist of 750 biopsy images of the human duodenum  which have a total of 6800 polyline annotations of Villi and Crypts. Additionally, each image is provided with annotated villi shoulder and crypt border. We evaluate the performance of \model{}  on CeDeM for localization, measurement and CeD detection. When compared to its closest baseline, \model{} obtains gain of 25\% in localization, 33\% in measurement, 9\% in binary classification and 11\% in grade classification accuracy. 

In summary: (a) We introduce the task of CeD detection through measurement, utilizing polyline detection instead of traditional methods, and demonstrate its advantages; (b) We propose \model{}, a polyline detection framework that incorporates localization and object-driven losses specifically designed for accurate measurement; (c) Guided by pathological insight, we use segmentation masks as auxiliary information to improve crypt length and enhance feature fusion robustness through mixup; and (d) We present CeDeM, a novel dataset of human duodenum biopsy images, containing 6,800 annotated villi and crypts, to support research in automated celiac disease detection.

\section{Related Work}

\begin{figure*}[t]
    \centering
    \includegraphics[scale = 0.5, width=17.6cm]{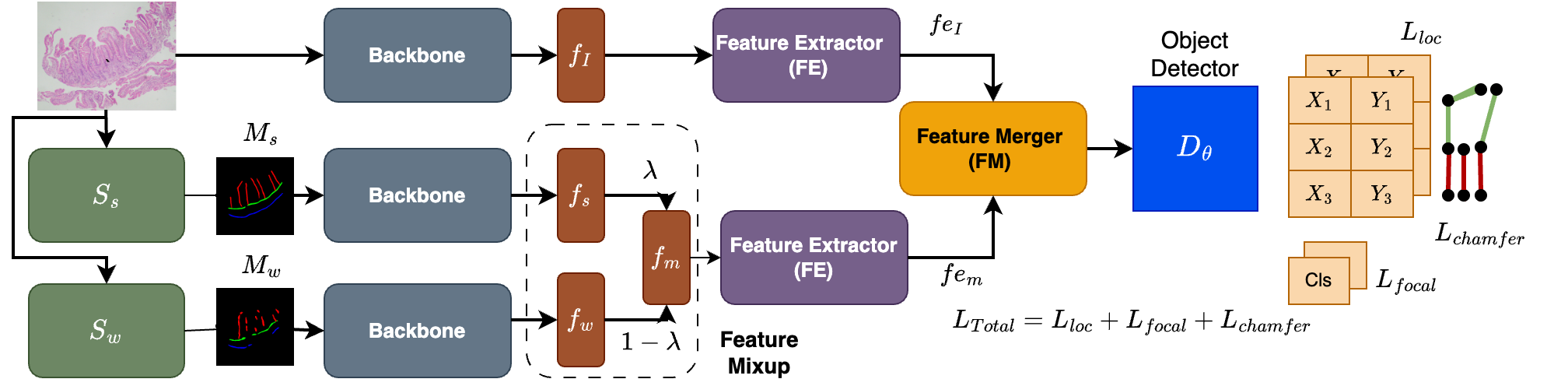} 
    \caption{
During training, given an input image ($I$), the segmentation models \( S_{s} \) and \( S_{w} \) predict strong (\( M_{s} \)) and weak (\( M_{w} \)) segmentation masks. Features are extracted from the image (\( f_{I} \)) as well as from each mask (\( f_{s} \) and \( f_{w} \)). To mitigate over-dependence on mask features and reduce exposure bias, we apply feature mixup (\( f_{m} \)) before the Feature Extraction (FE) stage. The image and mixed mask features are then combined using the Feature Merger (FM) module, producing a unified image-mask feature representation. These merged features are passed through the detection model \( D_{\theta} \), a polyline detection framework, which is trained using polyline detection and object-driven loss. During inference, \model{} utilizes only \( S_{s} \), eliminating the need for feature mixup.}

    \label{fig:Architecture}
\end{figure*}

\textbf{Celiac Disease:}
Pathologists rely on parameters like Q-histology scoring, which reflects increased intraepithelial lymphocytes (IEL), and villi-to-crypt length ratios, which indicate villous atrophy, to support CeD diagnosis \cite{das2019quantitative}. Most solutions for identifying CeD use classification approaches \cite{carreras2024celiac, puttmann2024automated, sali2019celiacnet}, which are often non-interpretable and offer limited assistance to pathologists. DeGPR \cite{tyagi2023degpr} applies the Q-histology score to classify CeD by counting IELs at villi tips but does not assess CeD severity. In this work, we focus on grading CeD through villi-crypt length measurement. 

\textbf{Segmentation-based Methods:} One approach to address length measurement problem is to treat it as a segmentation task, where for a predicted segmentation mask, post-processing is applied to extract length measurements. For applications like organ or tumor size measurement, U-Net \cite{ronneberger2015u} combined with EfficientNet \cite{tan2019efficientnet} is often used to estimate dimensions, though accuracy depends on segmentation quality \cite{sharbatdaran2022deep, koukoutegos2024segmentation, hsiao2022deep, cai2018quantitative}. Low-confidence regions in segmentation models can cause fragmented predictions, resulting in errors in final measurements.

\textbf{Lane Detection based methods}: Lane detection involves predicting points along lanes within an image. Though lane detection techniques have not been directly used for length measurement, they can be adapted to point prediction tasks, where fitting a line through predicted points can be used to get final line. Traditionally, lane detection has been addressed using segmentation-based methods, such as SpatialCNN \cite{pan2018spatial}. LaneCNN \cite{li2019line} and PolyLaneNet \cite{tabelini2021polylanenet} approach lane detection as a detection task, predicting line anchors or polynomial coefficients for lanes. Methods like LaneATT \cite{tabelini2021keep} use attention-based anchors, while YoLino \cite{meyer2021yolino} applies a single-shot detection. These methods perform well with structured, mostly straight lanes, but are less suited to biopsy images where villi have highly variable orientation and steeper curvature. Our experiments demonstrate the limitations of these methods in measurement.


\textbf{Line Segment detection}:
Line segment detection involves predicting line segments from an image. Traditional methods, like LSD \cite{von2008lsd}, rely on image gradients and heuristics, while recent deep learning approaches, such as Deep-LSDNet \cite{pautrat2023deeplsd} and LSD-NET \cite{teplyakov2022lsdnet}, use learnable features and post-processing to refine predictions. HAWP \cite{xue2020holistically} predicts both line endpoints and an attraction field, enabling mutual refinement.
These methods typically detect edges, junctions, or region proposals to compute line segments \cite{pautrat2023deeplsd, lin2024comprehensive}, but are not directly applicable to our task, which focuses on detecting objects (e.g., villi and crypts) rather than edges. A more recent approach, LETR \cite{xu2021line}, uses an end-to-end transformer architecture inspired by DETR \cite{carion2020end}, incorporating multi-scale features and L1 loss for endpoint detection. We compare LETR’s performance with our approach and demonstrate improved results.


\section{Method}

In the polyline detection task, given a histopathology image \( I \), the goal is to predict a polyline $P = \{P_1, P_2, \dots, P_n\}$ and its corresponding class labels $C = \{c_1, c_2, \dots, c_n\}$. Each polyline is defined by three points: the starting point \( P_s \), the middle point \( P_m \), and the endpoint \( P_e \), where each point is represented by its \( (x, y) \) coordinates. Predicted polylines are then used to compute the length of villi/crypts. 

As a possible solution to the aforementioned problem we propose \model{}, a transformer-based polyline detector. \model{} is based on DINO-DETR \cite{zhang2022dino}, denoted as $D_{\theta}$ (see Fig \ref{fig:Architecture}). DINO-DETR ($D_{\theta}$), originally designed for bounding box detection, is adapted to perform polyline detection. Given an image $I$, \model{} predicts a set of polylines $P$ with start, middle, and end points for each instance, along with its class label (villi/crypt in our case).  Detection model ($D_{\theta}$) predicts 6 coordinates for each polyline instance, representing the $(x, y)$ coordinates of the start, middle, and end points. Given the training data, the detection model ($D_{\theta}$) is trained with localization and classification losses. Additionally, to accurately capture the line's curvature, \model{} use the chamfer distance \cite{homayounfar2018hierarchical}, which computes the distance between ground truth and predicted polylines. For our primary objective—accurate measurement—we propose object-driven losses, including a length loss and a part-length loss. The length loss ensures alignment between the predicted and ground truth lengths, while the part-length loss ensures that distances between polyline segments (i.e., start-to-middle point and middle-to-end point) closely match those in the ground truth.

In duodenum biopsy, due to slicing error, the manner in which the tissue is cut often result in crypts being only partially visible (see Fig \ref{fig:dataset_images}). Pathologists make approximate inference from the global attributes like villi shoulder (VS) and crypt border (CB) where they approximate the crypt length as distance between villi shoulder and crypt border (see Fig \ref{fig:dataset_images}). We use this pathological insight in \model{} to improve the crypt prediction. Villi shoulder gives global picture about starting point of villi. Crypt border visually inferred as lower boundary of biopsy, intersect with end point of crypts. We train the segmentation model ($S_{\theta}$), to get segmentation mask for villi shoulder and crypt border. While training $S_{\theta}$, we apply dice and cross-entropy losses, along with a dynamic time warping loss to ensure that each villi shoulder projection aligns with a corresponding point on the crypt border. During training of \model{}, we keep segmentation model $S_{\theta}$ frozen and use the segmentation mask prediction for image $I$, as auxiliary information to improve the polyline detection. 


\model{} uses Feature Extraction (FE) and Feature Merger (FM) blocks to create a Unified-Image-Mask (UIM) representation, integrating both image and auxiliary mask features. The UIM is then passed to the detection model ($D_{\theta}$) to generate polylines. To prevent $D_{\theta}$ from exposure bias \cite{arora2022exposure}—an over-reliance on auxiliary mask information—\model{} employ both strong and weak segmentation models, $S_{s}$ and $S_{w}$, to obtain auxiliary segmentation masks $M_{s}$ and $M_{w}$. \model{} then apply feature mixup \cite{zhang2017mixup} on the extracted features $f_{s}$ and $f_{w}$, creating a robust mask feature representation, $f_{m}$, as shown in Fig. \ref{fig:Architecture}.

\subsection{Polyline Detection}
\model{}, is trained with localization loss $L_{loc}$, which computes the error between the predicted and ground truth coordinates. The localization loss, is defined in Eq. \ref{eq:localization_loss}, computes the L1 distance between each instance of the predicted polyline $\hat P_{i}$ and corresponding ground truth polyline $P_{i}$

\begin{equation}
L_{loc} = \sum_{i=1}^N ||P_{i} - \hat P_{i}||_1
\label{eq:localization_loss}
\end{equation}




While individual point localization is crucial, the overall structure of the polyline formed by these points is equally important, as the predicted line $\hat{P}_{i}$ should match the curvature of the ground truth polyline $P_{i}$. To ensure that the predicted polylines closely resemble the ground truth in shape, \model{} use the Chamfer distance (CD) \cite{homayounfar2018hierarchical} as a loss function. This helps minimize the distance between the predicted and ground truth lines, ensuring similar curvature. The Chamfer loss for the $i$-th polyline is given as $L_{CD}$, where $p_{i,j} \in P_{i}$ and $\hat{p}_{i,k} \in \hat{P}_{i}$, and is defined in Eq. \ref{eq:chamfer_distance}.

\begin{equation}
    L_{CD} (P, \hat{P}) = \frac{1}{|P|} \sum_{j} \min_{k} d(p_{j}, \hat{p_{k}}) + \frac{1}{|\hat P|}\sum_{k} \min_{j} d(p_{j}, \hat{p_{k}})
    \label{eq:chamfer_distance}
\end{equation}
where $d(p, \hat p)$ is defined as,
\begin{equation}
    d(p, \hat{p}) = \|p - \hat{p}\|_2^2
\end{equation}

Since there exists instances from multiple classes, with a class imbalance (more villi than crypts), \model{} employ Focal Loss \cite{8417976} as the classification loss ($L_{cls}$). Focal Loss helps mitigate the effect of class imbalance by down-weighting the loss assigned to well-classified examples, focusing more on hard-to-classify instances. For polyline $P_{i}$ with class label $y$ and predicted label $\hat{y}$ for $\hat{P}$, Focal Loss is defined in Eq. \ref{eq:focal_loss}:

\begin{equation} 
L_{focal}(y, \hat{y}) = - \alpha (1 - \hat{y})^\gamma y \log(\hat{y}) 
\label{eq:focal_loss} 
\end{equation}

For our primary objective—accurate measurement—we introduce object-driven losses: Length Loss ($L_{l}$) and Part-Length Loss ($L_{PL}$). Length Loss ($L_{L}$) ensures that the total length of the predicted polyline closely matches the ground truth length. For the $i$-th polyline $P_{i}$ with points $p$ represented by $(x, y)$ coordinates, the Polyline Length (M) is defined as defined as: Eq: \ref{eq:GT_length}, \ref{eq:Pred:Length}

\begin{equation}
    M (P) = \sum_{j=1}^2 \|p_{j} - p_{j-1}||_2^2 
    \label{eq:GT_length}
\end{equation}

\begin{equation}
     M (\hat P) = \sum_{j=1}^2 \|\hat p_{j} - \hat p_{j-1}||_2^2 
    \label{eq:Pred:Length}
\end{equation}

and Length loss $L_{l}$ for $i$-th polyline is defined in Eq \ref{eq:Length_Loss}:
\begin{equation}
    L_{L} = | M(P_{i}) - M(\hat P_{i})|
    \label{eq:Length_Loss}
\end{equation}




Similarly, the Part Length Loss ($L_{PL}$) ensures that the distances between the start and middle points, as well as between the middle and end points, are consistent with the ground truth. For polyline $P$ and $\hat P$ with start, middle and end points $p_{s}$, $p_{m}$, $p_{e}$ and $\hat p_{s}$, $\hat p_{m}$, $\hat p_{e}$ respectively, Part length loss ($L_{PL}$) loss is defined in Eq. \ref{eq:PLL_1}, \ref{eq:PLL_2}, and \ref{eq:PLL_combined}:

\begin{equation}
    L_{PLsm} = | d(p_{s}, p_{m}) - d(\hat p_{s}, \hat p_{m})|
    \label{eq:PLL_1}
\end{equation}

Here, \(d(p_{s}, p_{m})\) and \(d(p_{m}, p_{e})\) represent the distances between the start and middle points, and the middle and end points of the polyline, respectively.

\begin{equation}
L_{PLme} = |d(p_{m}, d_{e}) - d(\hat{p}_{m}, \hat{p}_{e})|
\label{eq:PLL_2}
\end{equation}

\begin{equation}
L_{PL} = L_{PLsm} + L_{PLme}
\label{eq:PLL_combined}
\end{equation}

Final loss is defined by Eq \ref{eq:final_loss}:
\begin{equation}
    L_{final} = L_{loc} + L_{CD} + L_{focal} + L_{L} + L_{PL}
    \label{eq:final_loss}
\end{equation}

\subsection{Mask Supervision}
Segmentation model $S(\theta)$ can be used to provide villi shoulder and crypt border. To achieve this, we train SegFormer \cite{xie2021segformer}, a end-to-end transformer-based segmentation model. Segmentation model is trained using a combination of Dice loss $L_{dice}$ and cross-entropy loss $L_{ce}$ to ensure robust segmentation. Dice loss optimizes the overlap between predicted and ground truth regions, and the cross-entropy loss ensures pixel-wise classification accuracy.



To ensure that each pixel in the villi shoulder (VS) has a corresponding projection onto the crypt border (CB), $S_{\theta}$ employ Dynamic Time Warping (DTW) ($L_{DTW}$) as loss function. We compute $L_{DTW}$ between predicted contour of VS and CB and loss in minimum when both are aligned. 





The total segmentation loss \( L_{\text{seg}} \) is defined as the combination of the Dice loss \( L_{\text{dice}} \), Cross-Entropy loss \( L_{\text{CE}} \), and Dynamic Time Warping loss \( L_{\text{DTW}} \):

\begin{equation}
L_{\text{seg}} = L_{\text{dice}} + L_{\text{CE}} + L_{\text{DTW}}
\label{eq:total_segmentation_loss}
\end{equation}

\subsection{Robust Mask Supervision}
Auxiliary segmentation masks can enhance the polyline detection performance; however, it may also lead to an over-reliance on these segmentation masks. During training, segmentation masks are typically highly accurate and closely aligned with the ground truth. In contrast, during the validation process, segmentation masks may contain errors, such as discontinuous predictions or false positives, which can result in inaccurate polyline predictions. To ensure robustness in the utilization of mask features, \model{} employ mixup \cite{zhang2017mixup} on the mask features.

To implement this, \model{} utilize two segmentation models: one trained to full capacity, referred to as the Strong Segmentor (\(S_{s}\)), and another trained to 50\% capacity, known as the Weak Segmentor (\(S_{w}\)). Segmentation masks from \(S_{s}\) and \(S_{w}\) are passed through the DINO backbone to compute the mask features \(f_{s}\) and \(f_{w}\), respectively. Mask features $f_{s}$ and $f_{w}$ are used to create a mixed feature representation using Eq \ref{eq:mixup}:

\begin{equation}
    f_{m} = \lambda \cdot f_{s} + (1 - \lambda) \cdot f_{w}
    \label{eq:mixup}
\end{equation}

Here, \(\lambda\) is sampled from a Beta distribution with parameters $\delta$. Mixup ensures that the \model{} learns to incorporate mask features with some noise, enabling it to extract meaningful information despite inaccuracies.

\subsection{Image-Mask feature Fusion}
\model{} uses a unified image-mask feature representation, denoted as $f_{UIM}$, as input to the $D(\theta)$. To obtain $f_{UIM}$, \model{} apply Feature Extraction (FE) and Feature Merger (FM) blocks. Feature Extractor (FE) block, use inverted residual block of MobileNetV2 \cite{sandler2018mobilenetv2}. For image feature $f_{I}$, extracted features $fe_{I}$ is given by Eq \ref{eq:FE} and similarly for $f_{m}$ we compute $fe_{m}$.
\begin{equation}
fe_{I} = \text{ReLU} \left( \text{DWConv}\left( \text{ReLU} \left( \text{Conv} \left( \text{Conv}(f_{I}) \right) \right) \right) \right) + f_{I}
\label{eq:FE}
\end{equation}
Feature Merge (FM) receives features from image $f_{I}$ and mask $f_{m}$ and create $f_{UIM}$. Feature Merger (FM), inspired by the MLP-Mixer architecture \cite{tolstikhin2021mlp}, concatenates features $f_{I}$ and $f_{m}$ along the channel dimension, followed by a depth-wise convolution that expands the channels to $4x$, enabling FM to learn richer feature representations. This expanded channel is then split into two parts. One part is passed through a GELU activation, and the resulting tensors interact through a hadamard product. This process yields $f_{UIM}$ with the same dimensionality as the original image features $f_{I}$.

\begin{figure*}[t]
\centering
\begin{tikzpicture}[scale=1.0, transform shape, picture format/.style={inner sep=0.5pt}]

    \node[picture format]                   (A1)   at (0,0)            {\includegraphics[width=1.4in, height=1.4in]{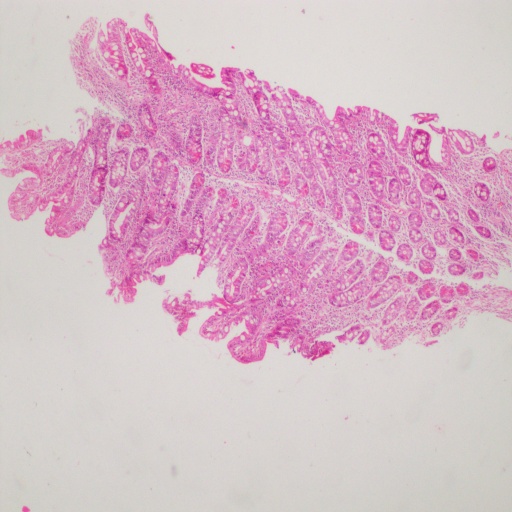}};
    \node[picture format,anchor=north]      (B1) at (A1.south)         {\includegraphics[width=1.4in, height=1.4in]{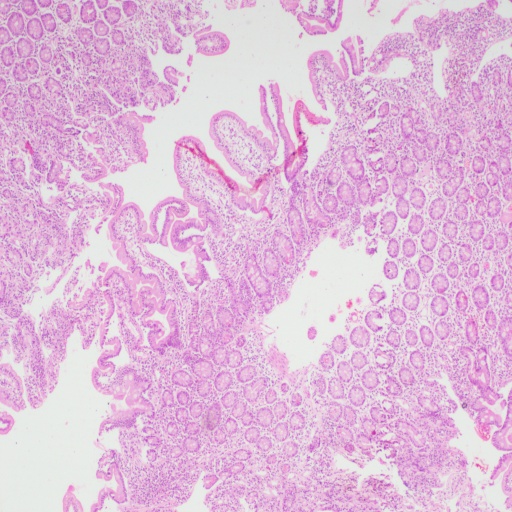}};
    \node[picture format,anchor=north]      (C1) at (B1.south)         {\includegraphics[width=1.4in, height=1.4in]{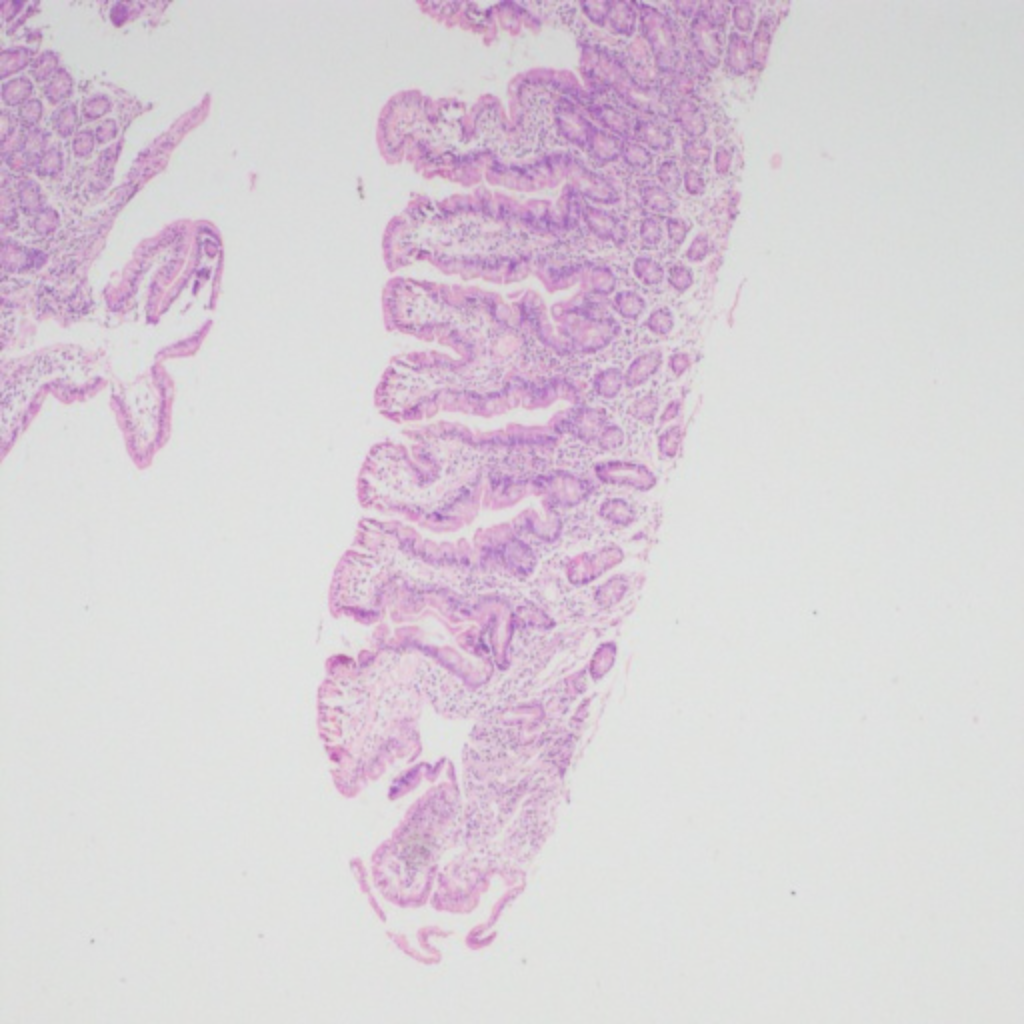}};

    \node[picture format,anchor=north west] (A2) at (A1.north east)    {\includegraphics[width=1.4in, height=1.4in]{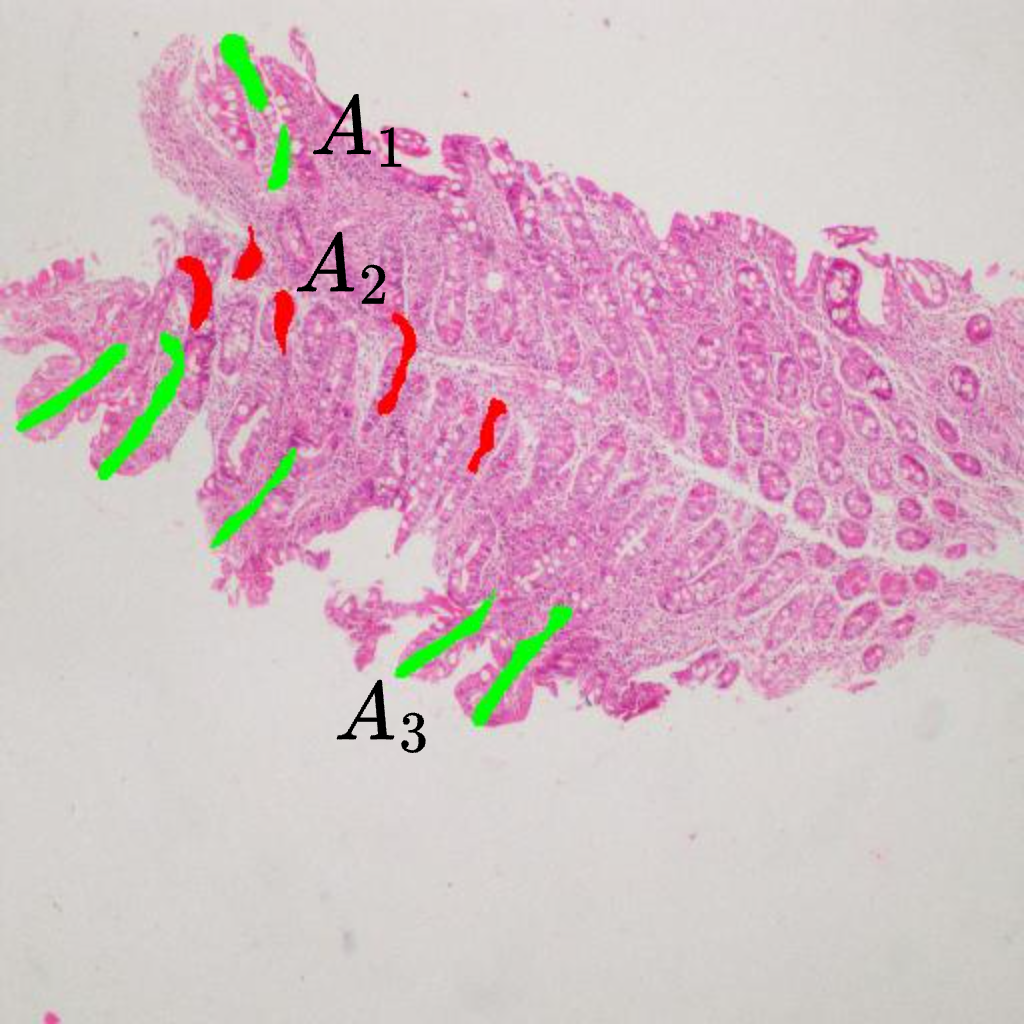}};
    \node[picture format,anchor=north]      (B2) at (A2.south)         {\includegraphics[width=1.4in, height=1.4in]{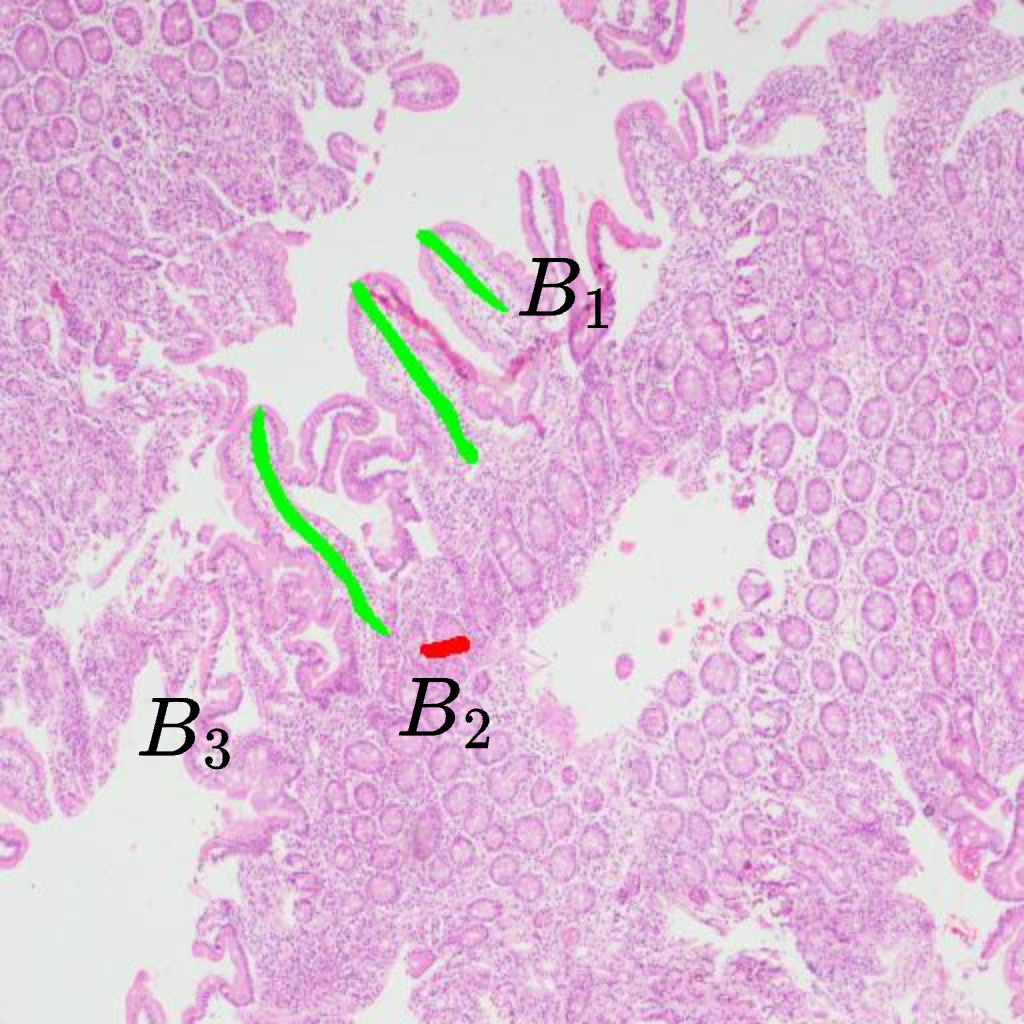}};
    \node[picture format,anchor=north]      (C2) at (B2.south)         {\includegraphics[width=1.4in, height=1.4in]{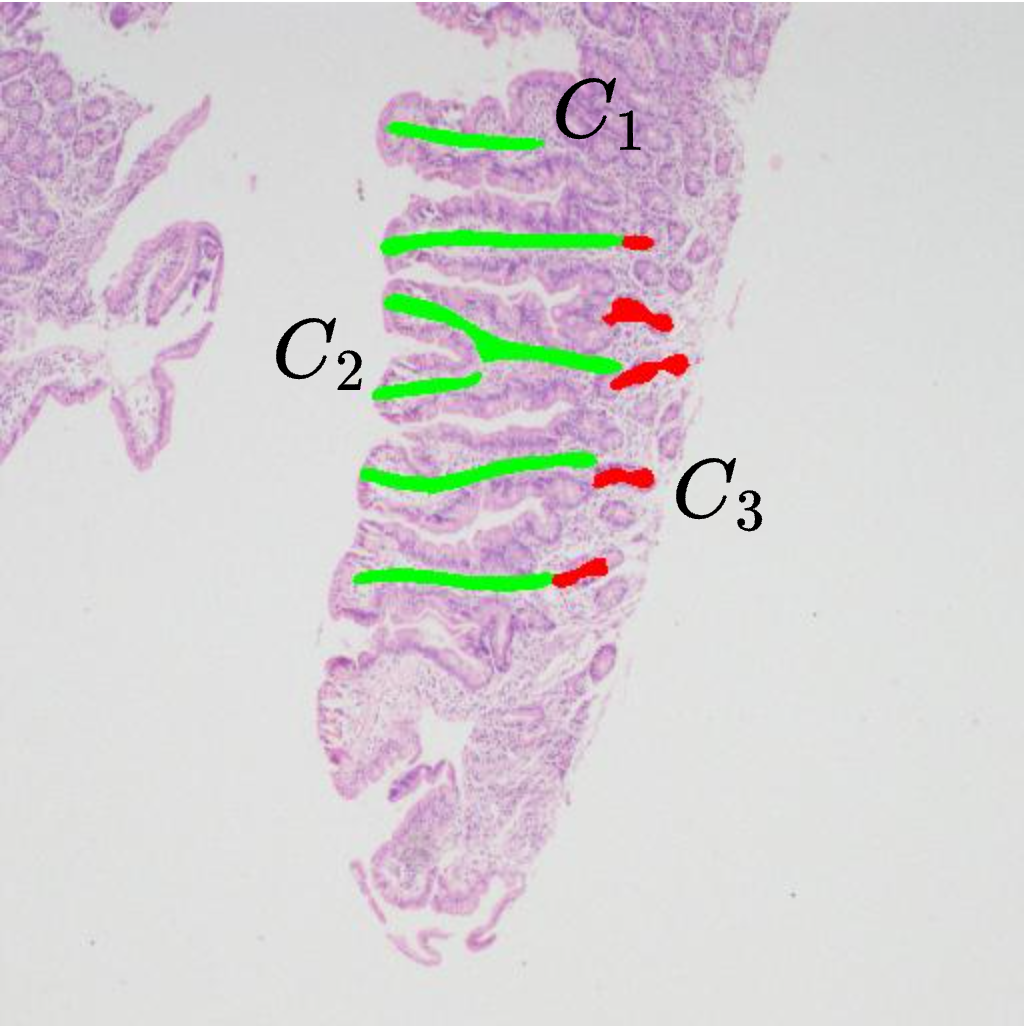}};

    \node[picture format,anchor=north west] (A3) at (A2.north east)    {\includegraphics[width=1.4in, height=1.4in]{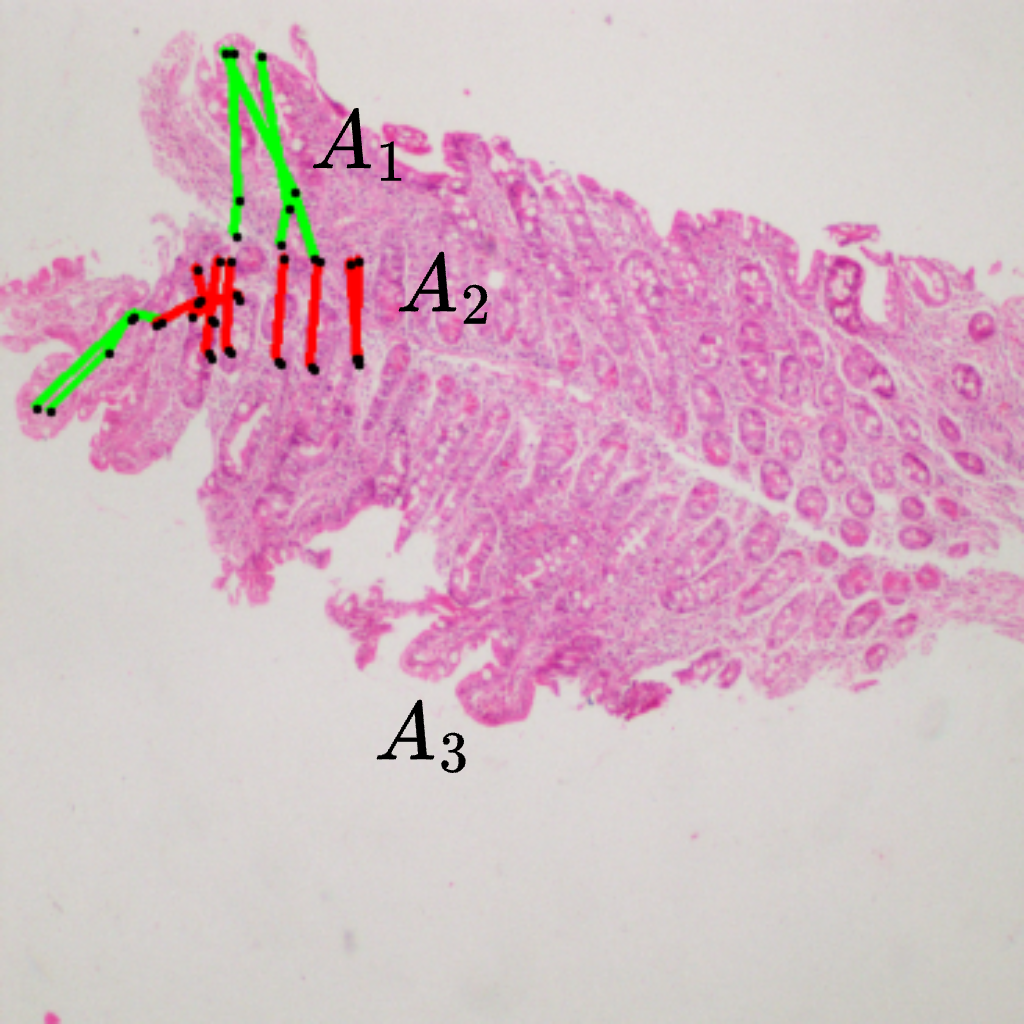}};
    \node[picture format,anchor=north]      (B3) at (A3.south)         {\includegraphics[width=1.4in, height=1.4in]{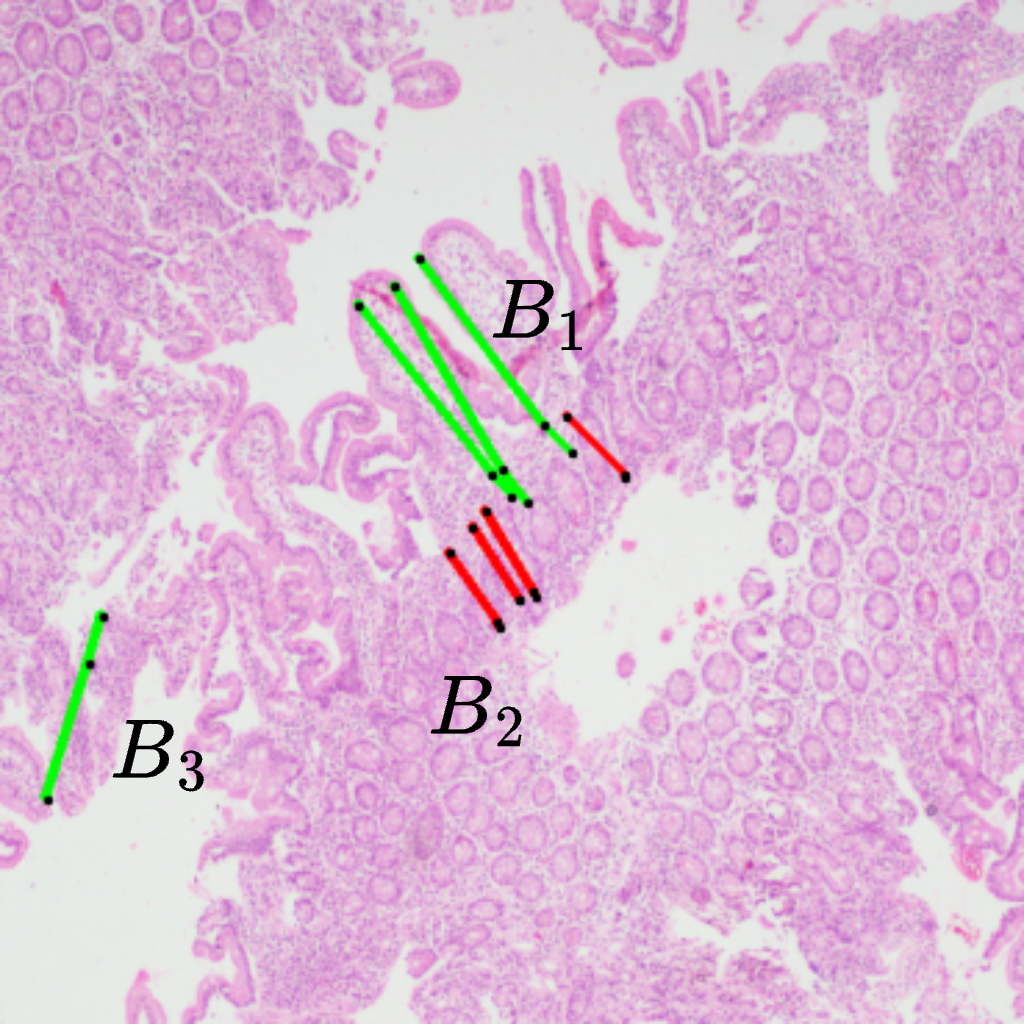}};
    \node[picture format,anchor=north]      (C3) at (B3.south)         {\includegraphics[width=1.4in, height=1.4in]{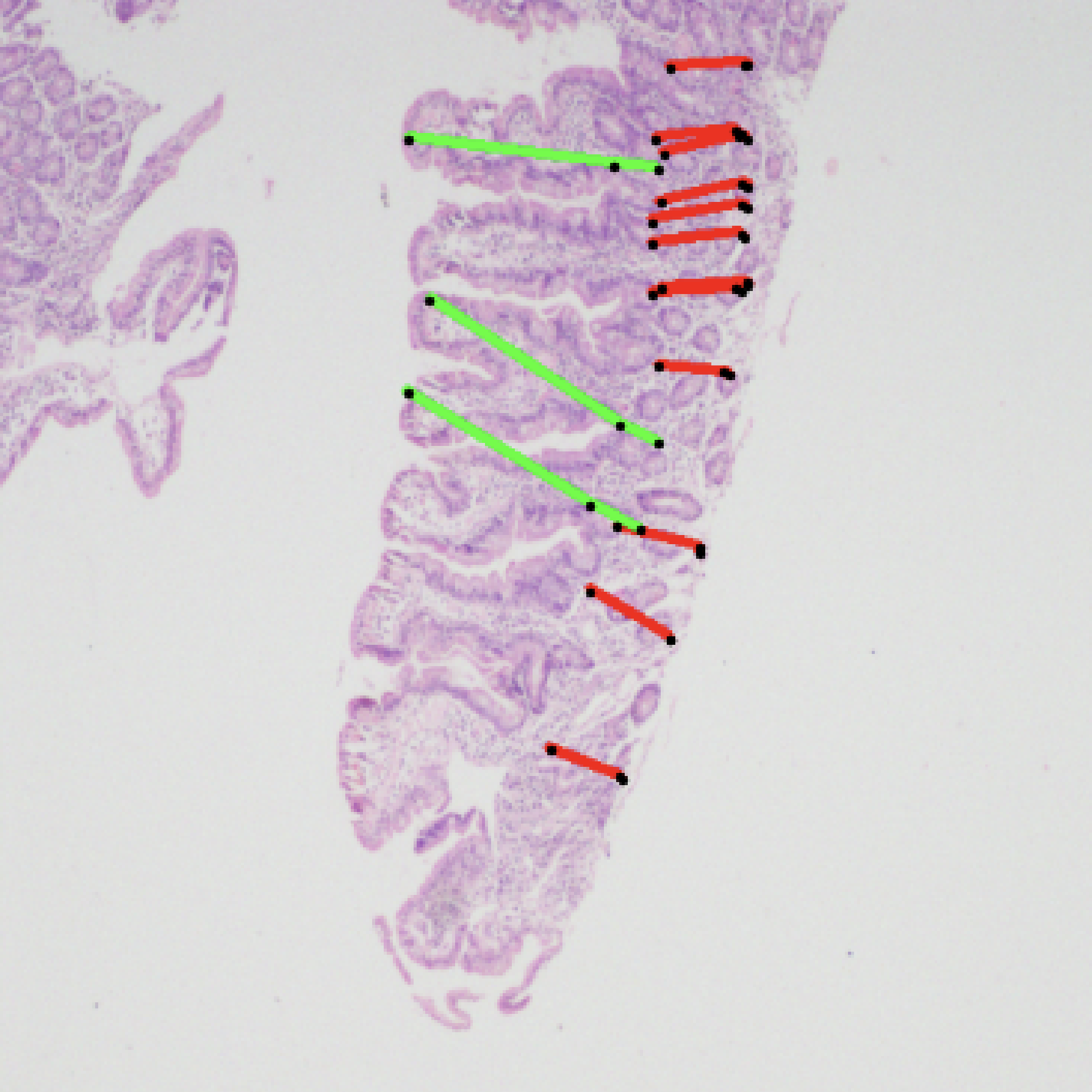}};

    \node[picture format,anchor=north west] (A4) at (A3.north east)    {\includegraphics[width=1.4in, height=1.4in]{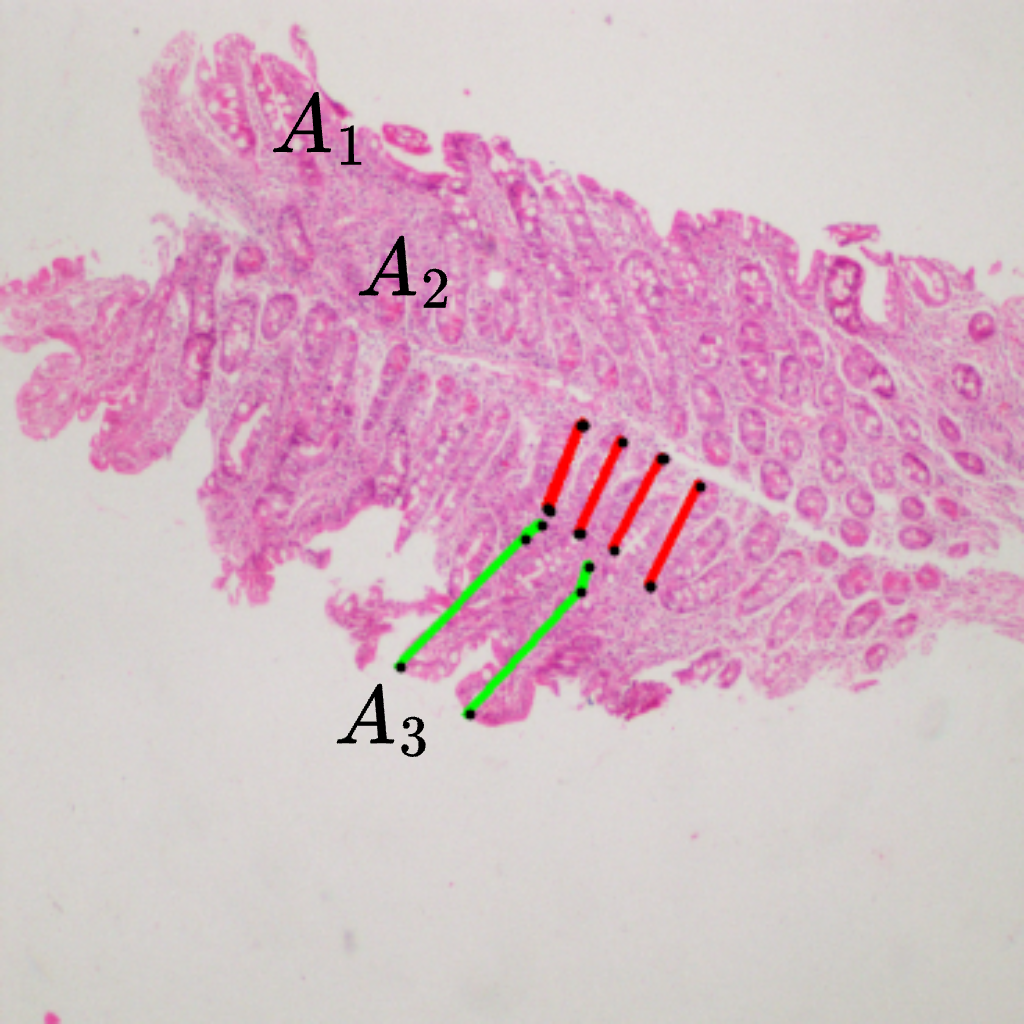}};
    \node[picture format,anchor=north]      (B4) at (A4.south)         {\includegraphics[width=1.4in, height=1.4in]{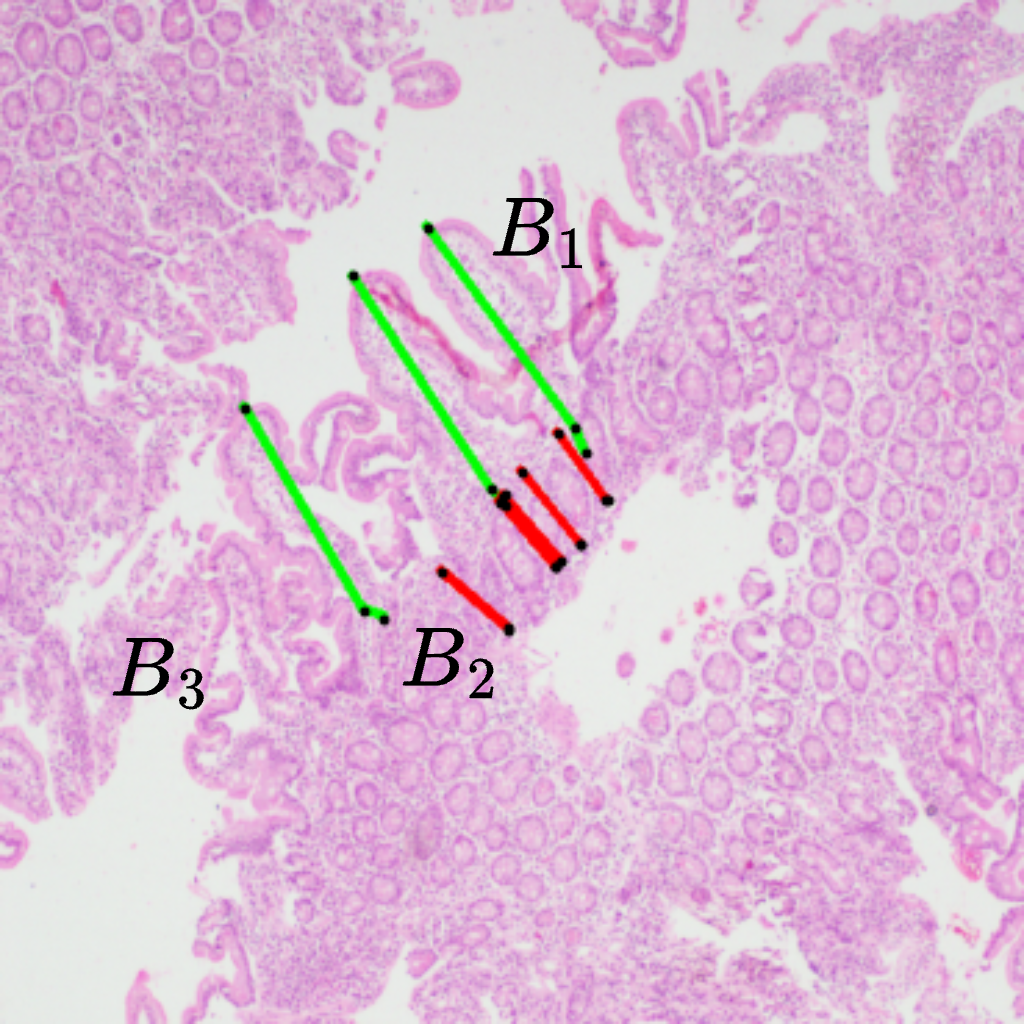}};
    \node[picture format,anchor=north]      (C4) at (B4.south)         {\includegraphics[width=1.4in, height=1.4in]{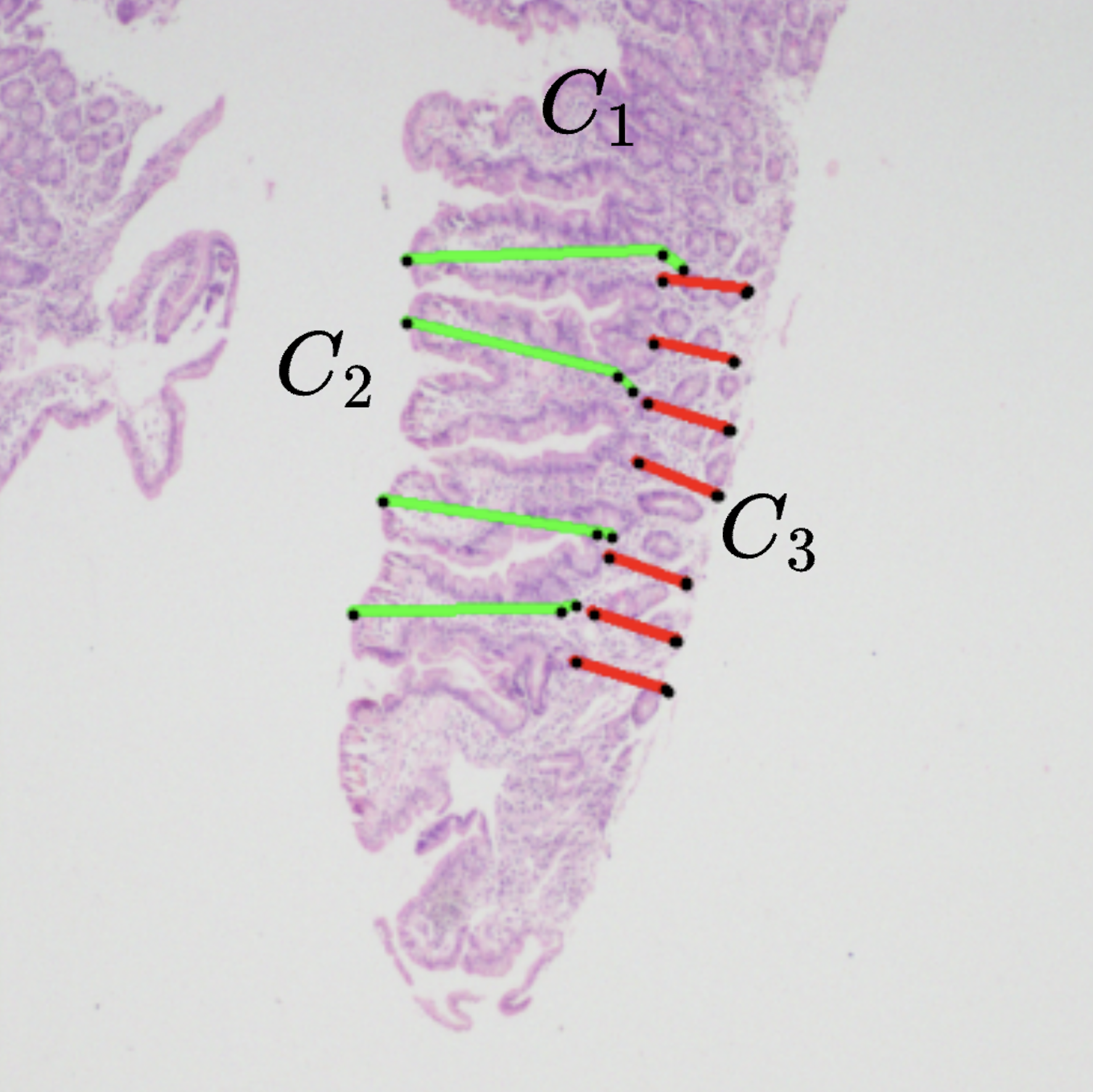}};

    \node[picture format,anchor=north west] (A5) at (A4.north east)    {\includegraphics[width=1.4in, height=1.4in]{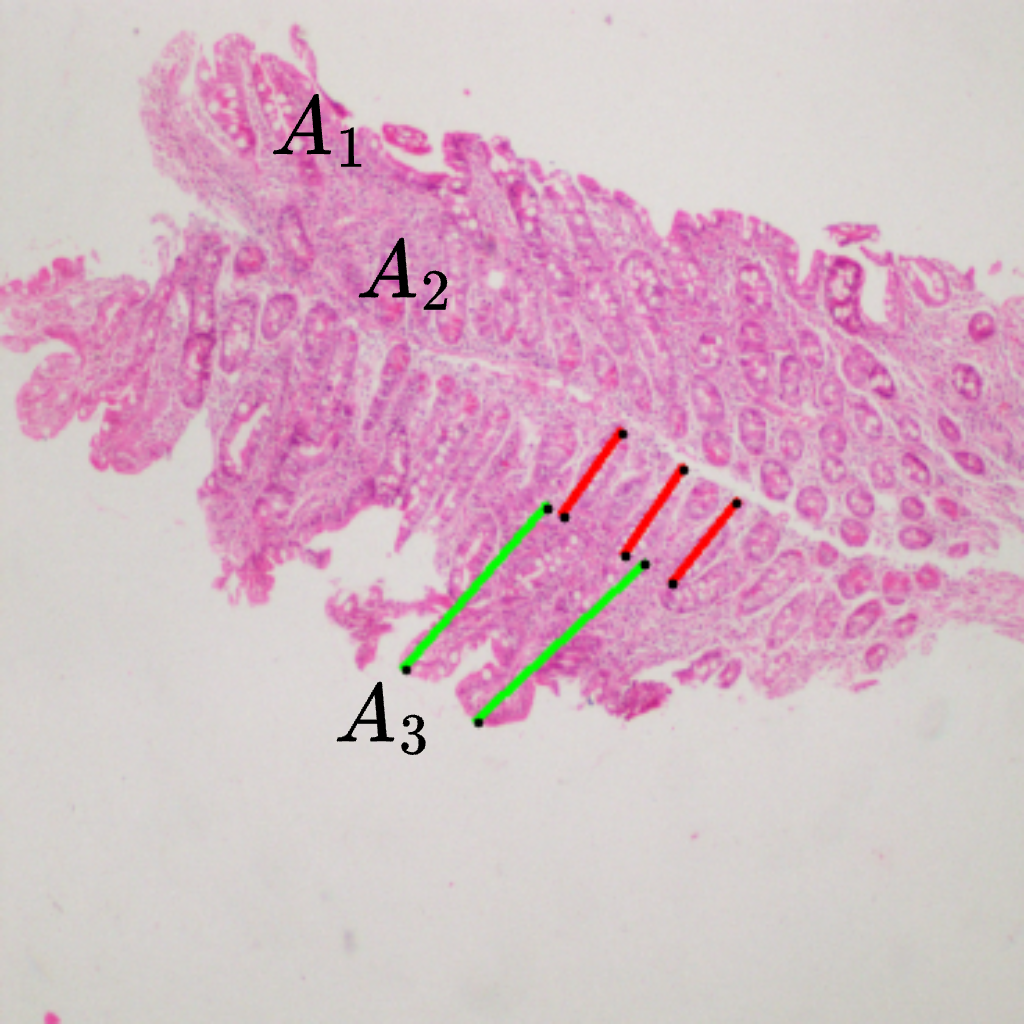}};
    \node[picture format,anchor=north]      (B5) at (A5.south)         {\includegraphics[width=1.4in, height=1.4in]{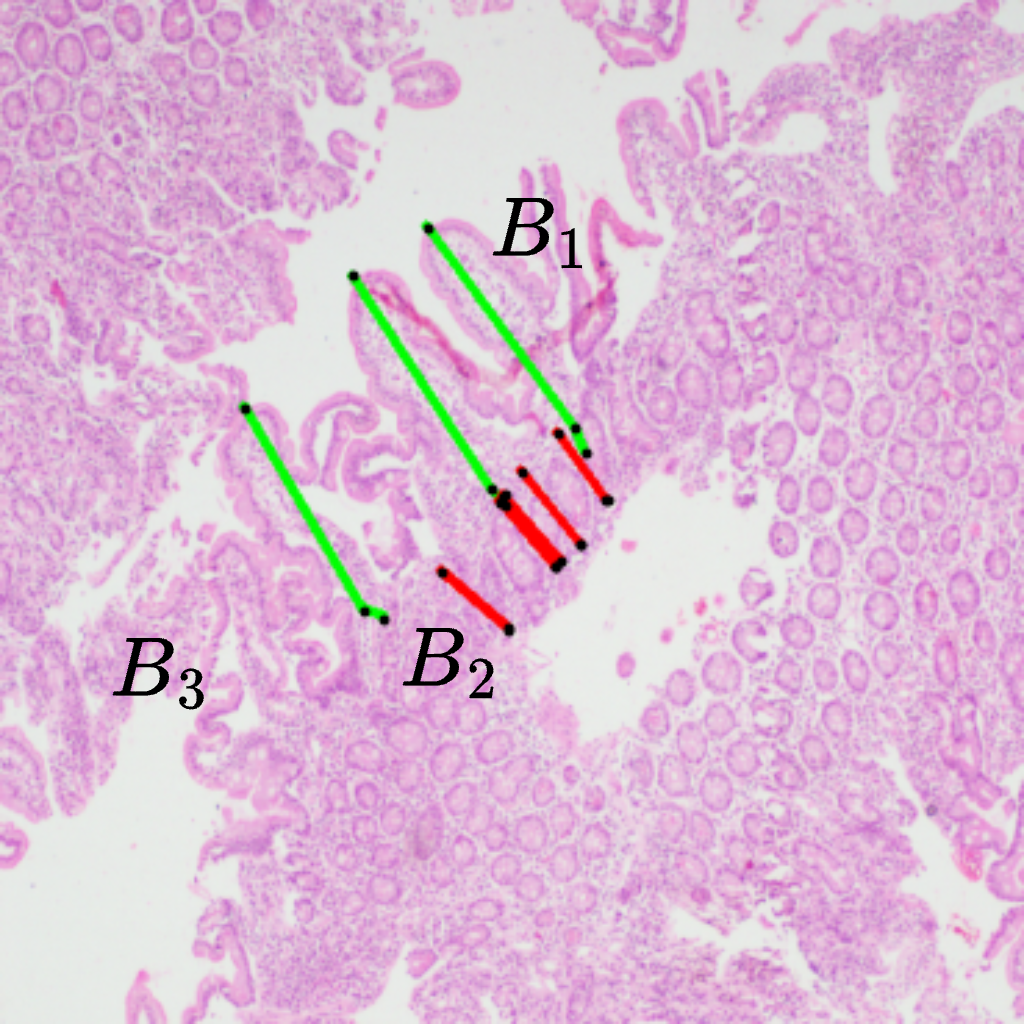}};
    \node[picture format,anchor=north]      (C5) at (B5.south)         {\includegraphics[width=1.4in, height=1.4in]{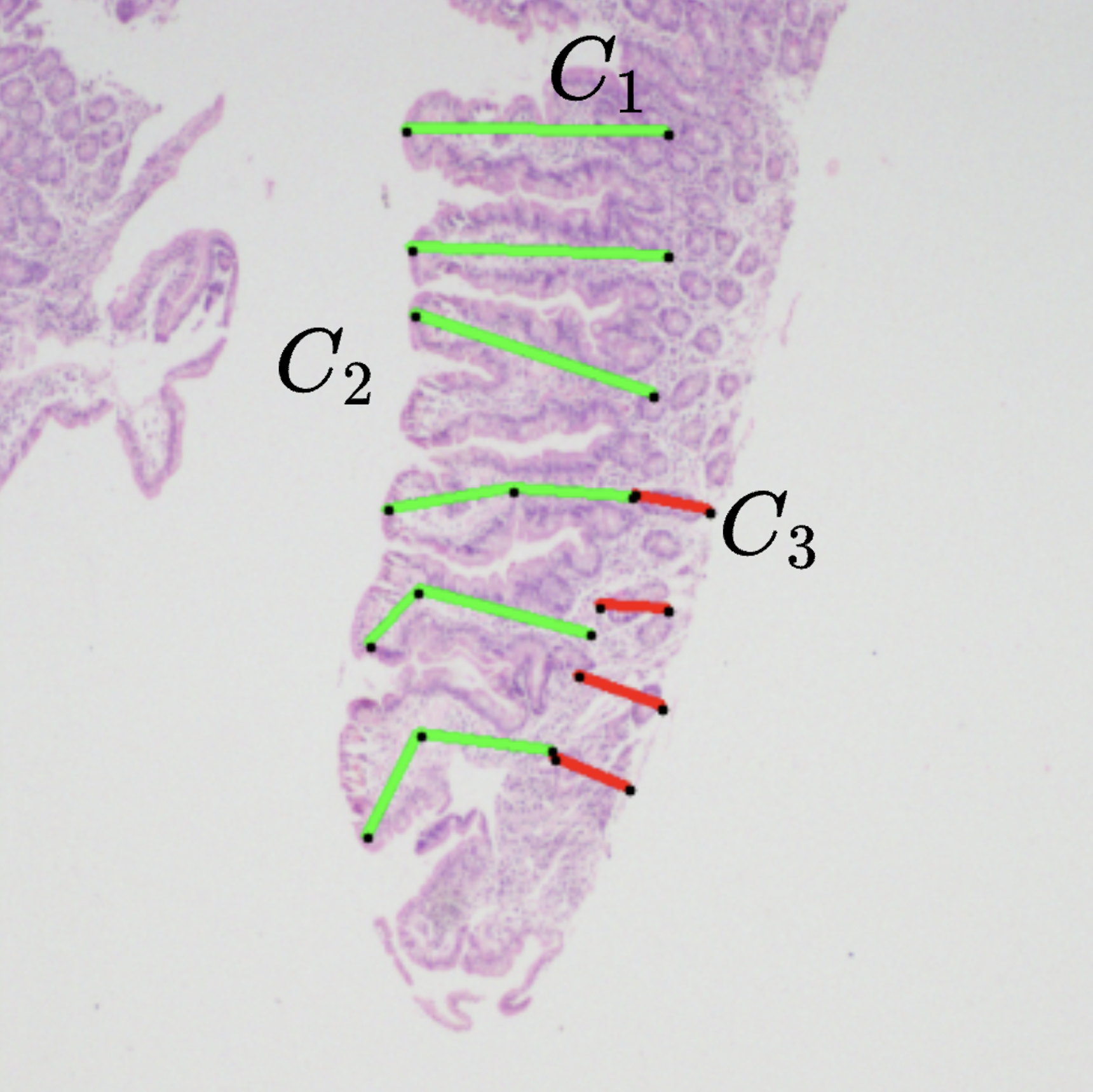}};

    \node[anchor=south] (N1) at (A1.north) {\bfseries Image};
    \node[anchor=south] (N2) at (A2.north) {\bfseries Segmentation};
    \node[anchor=south] (N3) at (A3.north) {\bfseries LETR};
    \node[anchor=south] (N4) at (A4.north) {\bfseries MeasureNet};
    \node[anchor=south] (N5) at (A5.north) {\bfseries Ground Truth};


\end{tikzpicture}
\caption{Qualitative performance of MeasureNet. First column shows Original Image, second, third, fourth and fifth column shows SegFormer, LETR, MeasureNet predictions and ground truth polyline respectively. Here, green and red corresponds to Villi and Crypts. The markers, $A_{1}$, $A_{2}$ and $B_{3}$ shows improvement in localization. Markers $A_{3}$, $B_{1}$, $B_{2}$ and $C_{1}$, shows improvement in partial predictions which cause error in measurement. Markers $C_{2}$ and $C_{3}$, shows better alignment and improved start-end points for villi and crypts.}
\label{fig:Detection_Visualization}
\end{figure*}
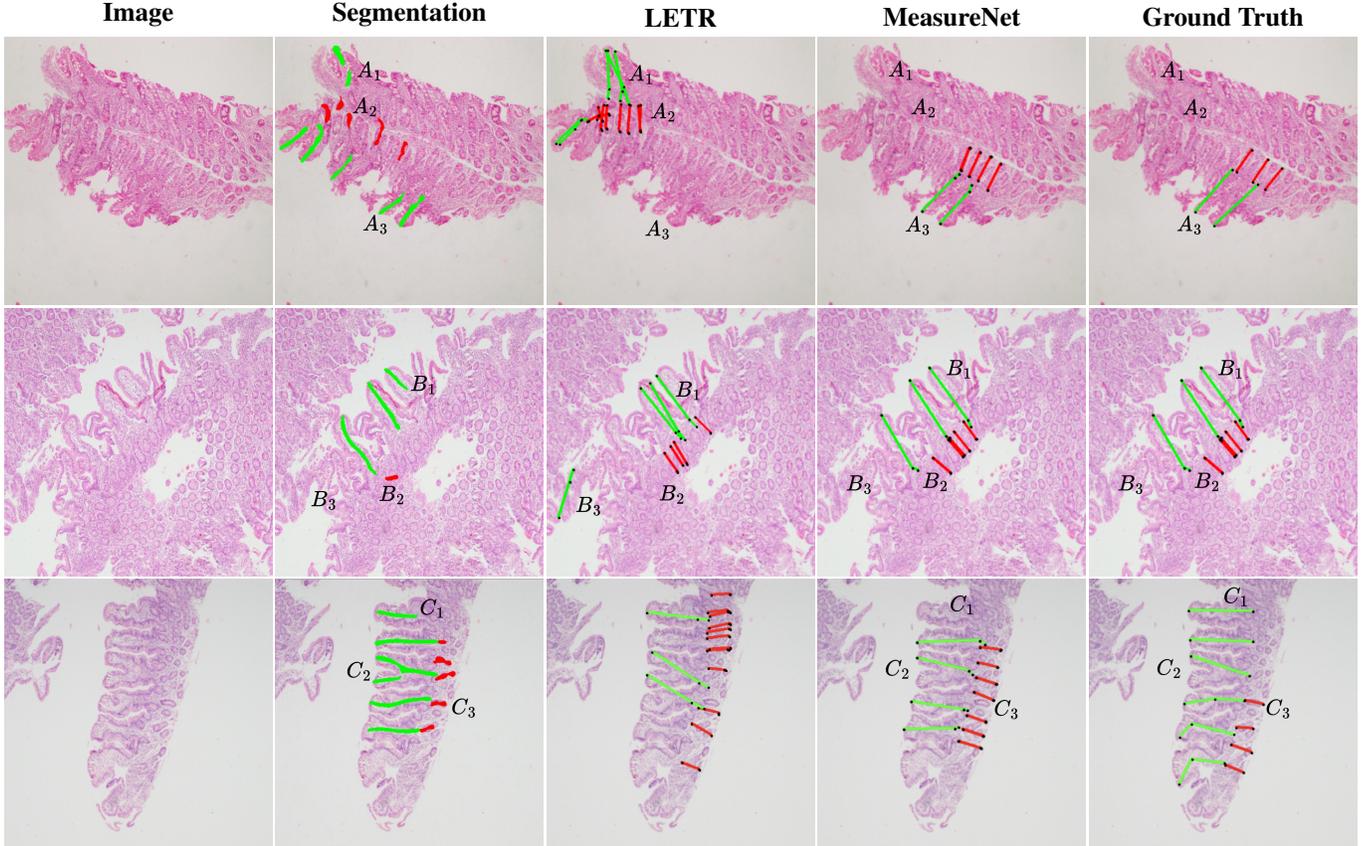

\section{Dataset Details}
We introduce a novel dataset, named Celiac disease Detection and grading through measurement (CeDeM), a dataset curated and validated by expert pathologists. CeDeM contains H\&E-stained histopathology images of the human duodenum, captured using an Olympus BX50 microscope at 4X magnification, with a DP26 camera, producing images of dimensions 1920x2148 pixels. CeDeM comprises 750 images, each with annotated polylines for both villi and crypts, amounting to 2617 annotated villi and 4211 annotated crypts. Additionally, each image consists of villi shoulder (VS) and crypt border (CB) annotations. All annotations were created using the LabelMe software and thoroughly validated by multiple pathologists. For training, dataset is divided into training, validation, and test sets, containing 610, 70, and 70 images, respectively. Each Villi/crypt is annotated with 2, 3, or 4 points with majority of them being 2 points. For \model{}, we converted every instance to 3 points, start, middle and end. For instances when only two points were available, we duplicated the start or end points to form a third point. For instances with more than three points, we sample one point between the start and end points.

\section{Experimental Setting}
Through our experiments, we aim to address the following research questions: (1) Can polyline detection be used for accurate length measurement? (2) What are the benefits of incorporating mask supervision in the polyline detection? (3) Does using mask mixup improve the model's robustness to noisy or imperfect masks? (4) What is the individual contribution of each component—such as polyline detection loss, object driven loss, mask and mask-mixup in enhancing polyline detection performance?

\subsection{Evaluation Metrics}
We evalaute the polyline detection method on 3 aspects, 1)  Localization performance, 2) Measurement 3) Celiac disease and Grade Classification. \\
\textbf{Localization:} We compute precision, recall and mAP to evalute localization performance. We measure true positives based on chamfer distance. Predicted polyline is considered as true positive if normalized chamfer distance between predicted polyline and ground truth line is below 0.05. For a ground truth polyline, if there is no predicted polyline with distance below 0.05, then we consider this as case of false negative. Similarly, for a predicted polyline if there is no ground truth polyline with distance below 0.05, then it is considered as false positive. 

\noindent\textbf{Measurement:} For evaluating measurement quality of the polyline, we compute Mean Absolute Error (MAE) and Mean Relative Error (MRE) for both villis and crypts. In addition to the individual measurements, we also compute the MAE and MRE for the ratio of villi to crypt length, based on the predicted polylines. MAE is the absolute difference between predicted and ground truth length/ratio and MRE measures the relative difference between predicted and ground truth lengths as a fraction of the ground truth.

\noindent\textbf{Celiac Classification:} From clinical perspective, the average villi length to crypt length ratio (Vd:Cd) plays a critical role in diagnosing celiac disease and assessing intestinal health. A Vd:Cd ratio greater than 3 indicates a normal patient, with healthy villi and crypts. If the Vd:Cd ratio falls between 1.05 and 3, the case is classified as Marsh Grade 1, indicating mild damage consistent with early celiac disease. A ratio between 0.95 and 1.05 corresponds to Marsh Grade 2, where villi and crypts are of nearly equal length, showing moderate damage. Finally, Vd:Cd ratio of less than 0.95, where the villi are shorter than the crypts, indicates Marsh Grade 3, the most severe stage, characterized by significant villous atrophy \cite{das2019quantitative}. This classification helps guide clinical decisions and treatment approaches for CeD. To evaluate the classification performance based on the measurements, we compute accuracy, precision, recall, and F1 score. 
Lastly, for evaluating performance the segmentation model ($S(\theta)$), we using the Dice coefficient and Intersection over Union (IoU).

\begin{table*}[ht]
\centering
\begin{tabular}{p{80pt} p{30pt} p{30pt} p{33pt} p{33pt}  p{30pt} p{30pt} p{40pt} p{33pt} p{30pt}}
\hline
  Model & MAE Villi $\downarrow$ & MRE Villi $\downarrow$ & MAE Crypt $\downarrow$ & MRE Crypt $\downarrow$ & MAE Ratio $\downarrow$ & MRE Ratio $\downarrow$ & Precision$\uparrow$ & Recall $\uparrow$ & mAP $\uparrow$  \\ \hline
  SegFormer \cite{xie2021segformer} & 31.63 & 29.41 & 17.76 & 36.61 & 0.858 & 0.371 & 36.62 & 55.72 & 40.91 \\
  BEiT \cite{bao2021beit} & 69.04 & 64.02 & 20.86 & 44.10  & 1.10 & 0.422 & 37.91 & 26.88 & 47.71 \\
  Yolino \cite{meyer2021yolino} & 69.76 & 41.54 & 25.12 & 30.59 & 1.65 & 0.788 & 14.30 & 54.19 & 20.31 \\
  LETR \cite{xu2021line} & 14.84 & 14.14 & 11.52 & 22.69 & 0.704 & 0.332 & 39.52 & 70.85 & 40.39 \\
  \hdashline
  MeasureNet (ours)& \textbf{14.39} & \textbf{13.59} & \textbf{6.83} & \textbf{13.87} & \textbf{0.4722} & \textbf{0.2116} & \textbf{51.37} & \textbf{79.94} & \textbf{50.29} \\ \hline
\end{tabular}
\caption{Measurement and detection results for CeDeM}
\label{tab:detection_table}
\end{table*}

\subsection{Implementation Details}
\model{} is build on top of the DINO-DETR. Specifically, we use the DETR model with 6 encoder and 6 decoder layers, leveraging the Swin Transformer \cite{liu2022swin} as the backbone. Detection model ($D_{\theta}$) is trained with learning rate of $1 \times 10^{-4}$, batch size 8, and query dimension 4. The input images are resized to $640 \times 640$ resolution. Sinusoidal positional embeddings are used to maintain spatial information throughout the model. We update the Bipartite Matching with chamfer distance. To enhance the model's robustness, we apply random augmentations, including rotations of 0°, 90°, 180°, and 270°, as well as horizontal, vertical, and diagonal flips. For $L_{focal}$, $\alpha$ is set as 0.25 and $\gamma$ is set as 2. For the segmentation task, we employ SegFormer \cite{xie2021segformer} with the B5 backbone, training the model for 250 epochs with a batch size of 8 and a learning rate of $6 \times 10^{-5}$. The images are resized to $640 \times 640$, and we apply augmentations similar to those used in the polyline detection model, including rotations of 0°, 90°, 180°, 270°, as well as horizontal, vertical, and diagonal flips. While training \model{}, we use mixup where $\lambda$ is sampled from beta distribution where $\delta$ is between [0.2, 0.4]. We did grid search on validation set to find the optimal hyper-parameters. All models are trained on an NVIDIA A100 GPU.

\section{Results}


\begin{table}[ht]
\centering
\begin{tabular}{p{45pt} p{25pt} p{40pt} p{25pt} p{40pt}}
    \hline
    & \multicolumn{2}{c}{Binary} & \multicolumn{2}{c}{Grade} \\ \hline
    & LETR & MeasureNet & LETR & MeasureNet \\ \hline
    Precision & 81.03 & \textbf{82.81} & 33.75 & \textbf{83.60} \\ 
    Recall    & 87.03 & \textbf{98.14} & 34.17 & \textbf{74.32} \\ 
    F1-score  & 83.92 & \textbf{89.83} & 33.33 & \textbf{75.61} \\ 
    Accuracy  & 73.91 & \textbf{82.66} & 70.10 & \textbf{81.42} \\ \hline
\end{tabular}
\caption{Metrics Comparison for Celiac Classification}
\label{tab:classification}
\end{table}

\begin{table}[ht]
\centering
\begin{tabular}{p{23pt} p{22pt} p{22pt} p{22pt} p{22pt} p{22pt} p{22pt}}
\hline
  Points & MAE Villi & MRE Villi & MAE Crypt & MRE Crypt & MAE Ratio & MRE Ratio \\ \hline
  2  & 17.99 & 15.79 & 9.03 & 17.27 & 0.6764 & 0.3019 \\
  3  & 14.39 & 13.59 & 6.83 & 13.87 & 0.4722 & 0.2116 \\
  4  & 17.35 & 15.52 & 8.46 & 16.82 & 0.6656 & 0.2758 \\
  \hline
\end{tabular}
\caption{Effect of number of points for polyline.}
\label{tab:number_of_points}
\end{table}

\begin{table*}[ht]
    \centering
    \begin{tabular}{p{30pt} p{30pt} p{30pt} p{30pt} p{30pt} p{25pt} p{25pt} p{25pt} p{30pt} p{30pt} p{30pt} p{30pt}}
    \hline
    Baseline & Polyline Loss & Length Loss & Part Length Loss & Mask & Mask Mixup & MAE Villi & MRE Villi & MAE Crypt & MRE Crypt & MAE Ratio & MRE Ratio\\\hline
    \checkmark & & & & & & 15.80 & 14.24 & 11.82 & 23.90 & 0.7140 & 0.3389\\
    \checkmark & \checkmark & & & & &14.84 & 14.14 & 11.52 & 22.69 & 0.7045 & 0.3329 \\
    \checkmark & \checkmark & \checkmark & & & & 14.94 & 14.70 & 10.09 & 20.33& 0.5749 & 0.2650\\
    \checkmark & \checkmark & \checkmark & \checkmark & & & \textbf{12.66} & \textbf{11.65} & 9.76 & 18.38 & 0.5480 & 0.2501\\
    \checkmark & \checkmark & \checkmark & \checkmark & \checkmark & & 16.98 & 15.55 & 7.01 & 13.89 & 0.5134 & 0.2225 \\
    \checkmark & \checkmark & \checkmark & \checkmark & \checkmark & \checkmark & 14.39 & 13.59 & \textbf{6.83} & \textbf{13.87} & \textbf{0.4722} & \textbf{0.2116} \\ \hline
    \end{tabular}
\caption{Ablation Table}
\label{tab:ablation}
\end{table*}

\subsection{Polyline Detection}

Table \ref{tab:detection_table} compares \model{} with several baseline methods, including segmentation-based approaches like SegFormer \cite{xie2021segformer} and BEiT \cite{bao2021beit}. 
These segmentation model predict segmentation masks for villi and crypt, from which we identify contours followed by skeletonize of contours to compute final lengths. We observe that \model{} obtains 38 pt reduction in MAE ratio (lower is better) and 10 pt mAP improvement as compared to Segmentation based methods. Segmentation based methods rely on visual attributes of villi/crypts to make prediction, however, these attributes can be mis-leading in some case. We also compare against Yolino \cite{meyer2021yolino}, a lane detection model. However, we observe that anchor-based approaches like Yolino struggle to capture the complex curvatures present in polylines. Next, we adapt LETR \cite{xu2021line}, originally designed for line prediction with two points, by integrating a Swin Transformer \cite{liu2022swin} backbone and modifying the decoder to predict three points instead of two. As compared to LETR, \model{} obtains 23 pt error reduction in MAE ratio and 10 pt mAP improvement. LETR suffers in overall measurement by missing the polyline based loss, object driven losses and auxiliary mask . It suggest that global attributes like villi shoulder and crypt border provide guidance to \model{} to obtain accurate measurement.


Qualitatively, we illustrate \model{} predictions in Fig \ref{fig:Detection_Visualization}. The first column shows the original image, the second column shows predictions from the segmentation model (SegFormer), third column shows images with prediction from LETR, fourth column shows predictions from \model{} and final column shows ground truth polylines. We observe that \model{} improve the localization performance as shown by highlighted markers $A_{1}$, $A_{2}$ and $B_{3}$, eliminate partial predictions shown by $A_{3}$, $B_{1}$, $B_{2}$ and $C_{1}$. Partial predictions hurt model performance by affecting the measurement eventually resulting in mis-classification. Finally, \model{} provide better alignment of predicted polyline with villi, shown by $C_{2}$ and $C_{3}$. 

Table \ref{tab:classification} reports classification performance based on villi-crypt-length ratio. \model{} improves CeD classification accuracy from 73.91\% to 82.66\% for binary classification task and CeD grading accuracy from 70.10\% to 81.42\% for multi-class classification task.

    

\subsection{Ablation Studies}
We perform ablation analysis to understand the relative contribution of different components in final performance, shown in Table \ref{tab:ablation}. Starting with DINO-DETR as the baseline (row 1) and adding Chamfer distance (row 2), we observe an improvement in MAE ratio by 0.014 points. Introducing length loss (row 3) further reduces the MAE ratio from 0.7045 to 0.5749. Adding part length loss (row 4) brings the ratio down to 0.5480, attributed mainly to improved curvature alignment in villi polylines. Incorporating mask information (row 5) significantly improves the MAE for crypts, from 9.76 to 7.01, emphasizing the auxiliary mask’s role in providing additional cues about villi shoulders and crypt borders. However, reliance on the segmentation mask can lead to false positives if the mask itself contains inaccuracies. To address this, we introduced mask mixup, which enhances crypt length estimation and improves robustness to false positives in the segmentation mask. Compared to row 5, \model{} with mask mixup achieves an error reduction of 2.59 points in villi MAE, 0.18 points in crypt MAE, and 0.04 points in the ratio MAE, demonstrating  gains in accurate measurement.

We analyzed the impact of varying the number of points used to represent a polyline for villi and crypt structures using our best-performing \model{}. Table \ref{tab:number_of_points} shows the measurement errors for different configurations. During training \model{}, we used a 3-point polyline representation. We evaluated \model{} trained with 2, 3, and 4 points against the original annotations. The results reveal an improvement of 3.9 points in measurement accuracy when increasing from a 2-point to a 3-point polyline. However, moving from a 3-point to a 4-point polyline showed negligible additional benefit. This indicates that a 3-point polyline is sufficient to capturing the villi-crypt structure.


\subsection{Error Analysis}

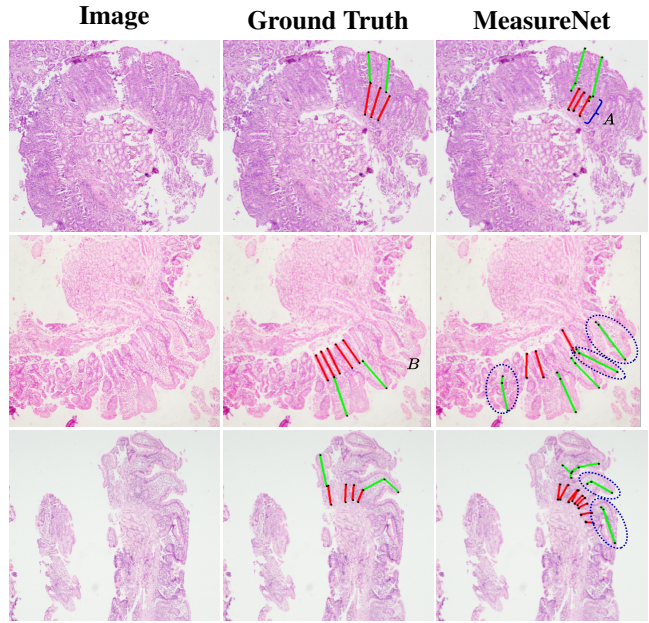
\begin{figure}[ht]
\centering
\begin{tikzpicture}[scale=1.0,transform shape, picture format/.style={inner sep=0.5pt}]

  \node[picture format]                   (A0)   at (0,0)            {\includegraphics[width=1.1in, height=1in]{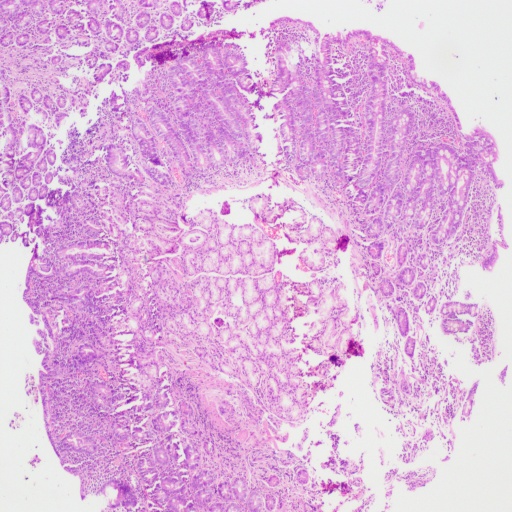}};
  \node[picture format,anchor=north west] (A1)   at (A0.north east)  {\includegraphics[width=1.1in, height=1in]{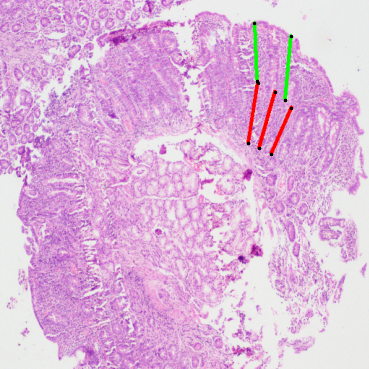}};
  \node[picture format,anchor=north west] (A2)   at (A1.north east)  {\includegraphics[width=1.1in, height=1in]{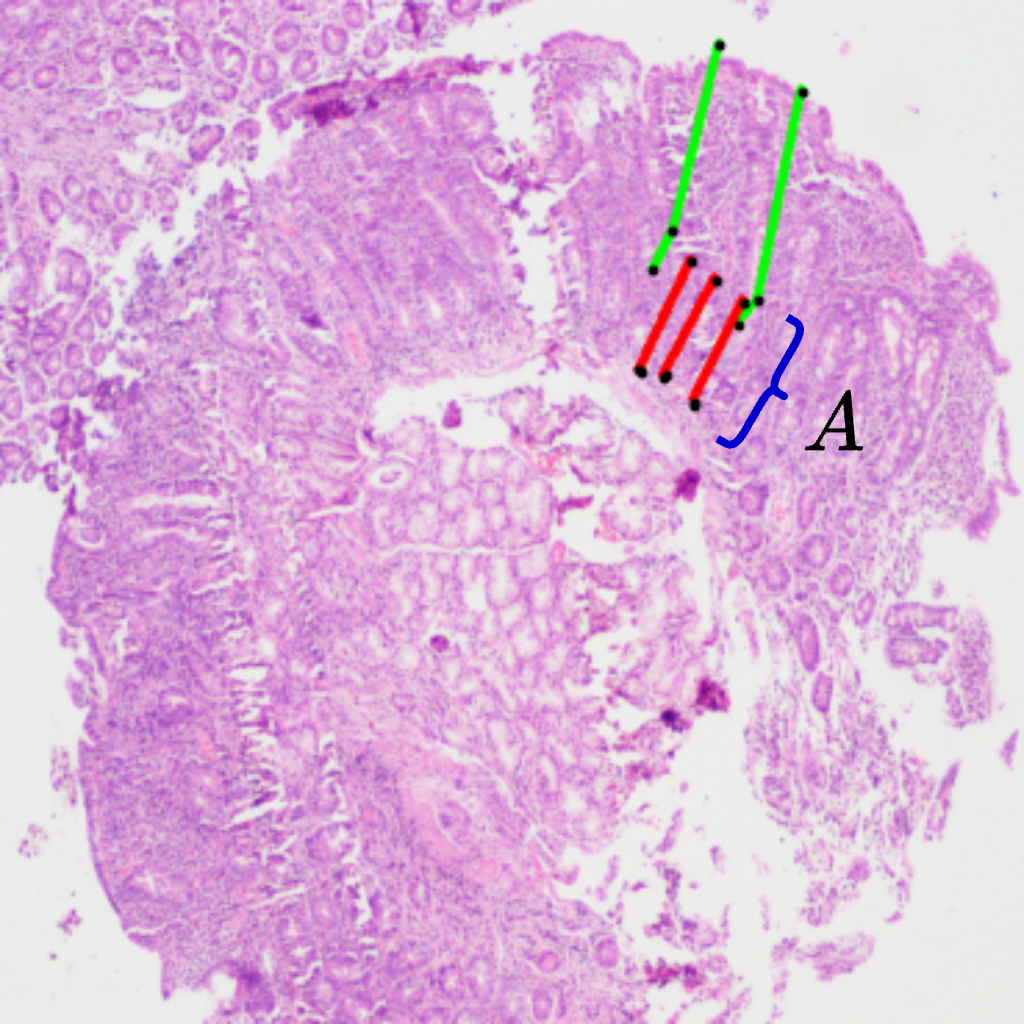}};

  \node[picture format,anchor=north]      (B0)   at (A0.south)       {\includegraphics[width=1.1in, height=1in]{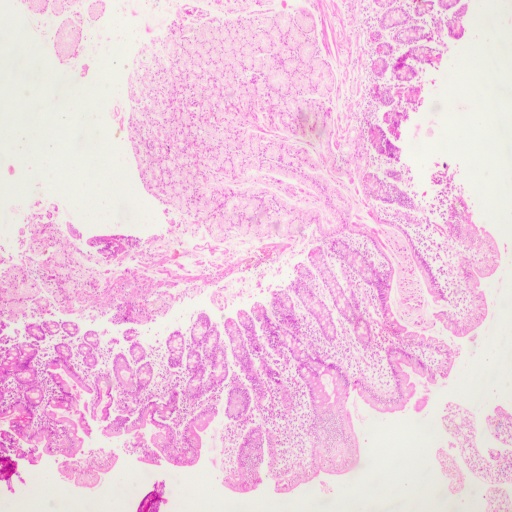}};
  \node[picture format,anchor=north]      (B1)   at (A1.south)       {\includegraphics[width=1.1in, height=1in]{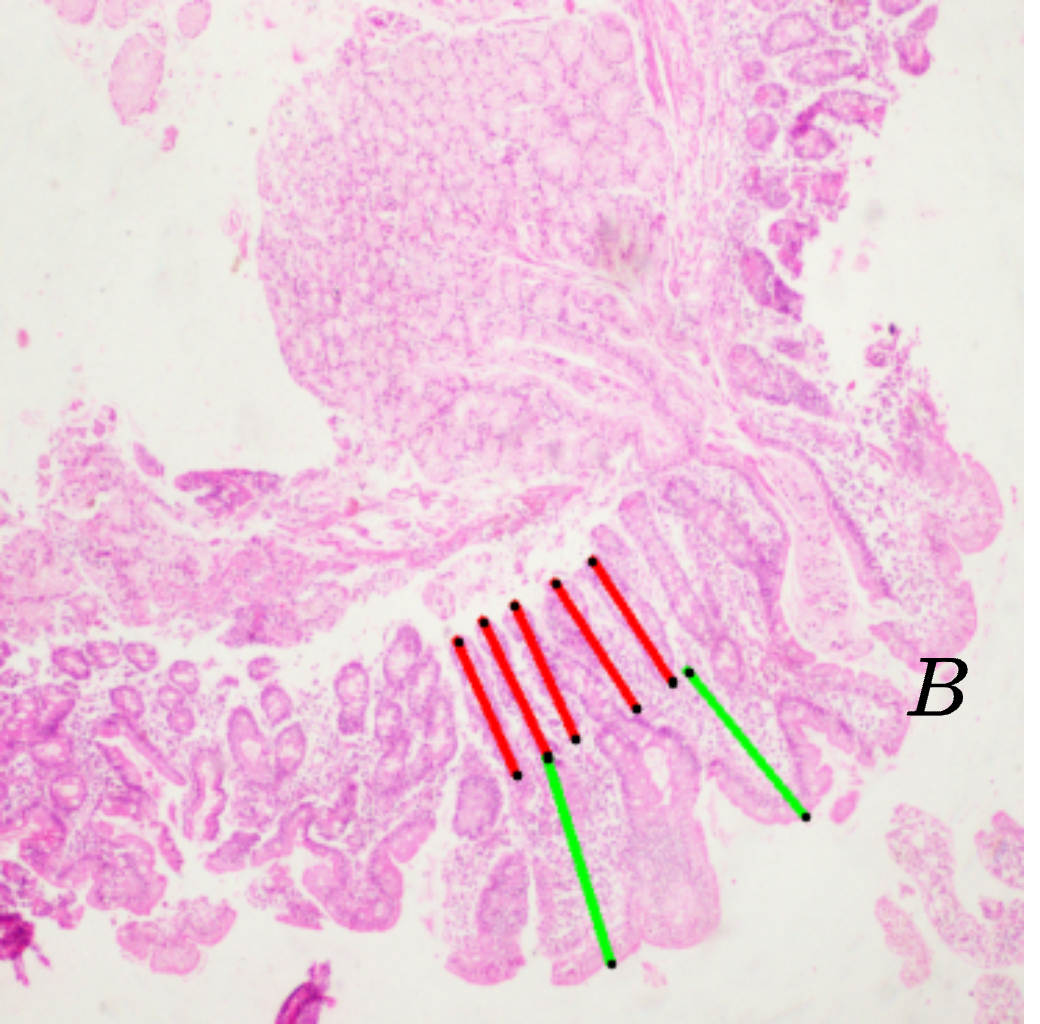}};
  \node[picture format,anchor=north]      (B2)   at (A2.south)       {\includegraphics[width=1.1in, height=1in]{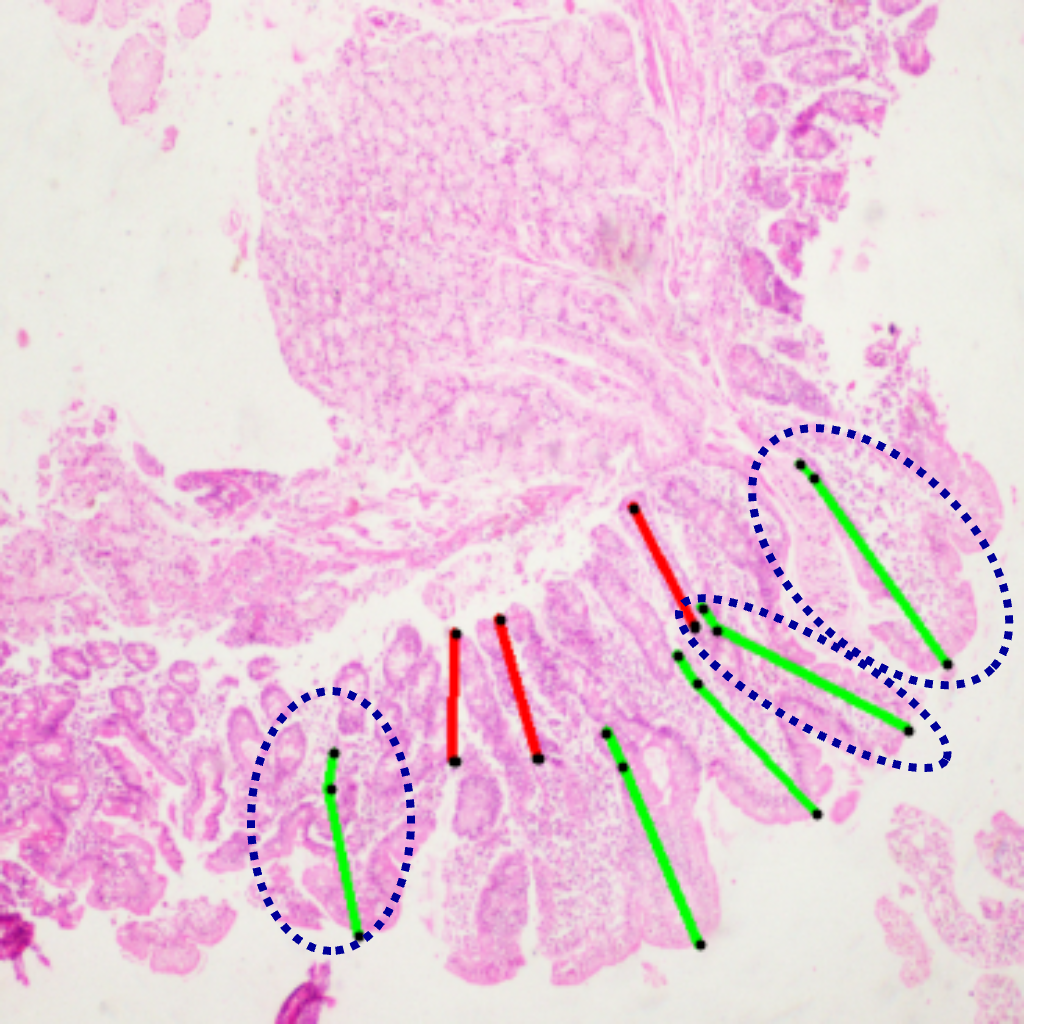}};

  \node[picture format,anchor=north]      (C0)   at (B0.south)       {\includegraphics[width=1.1in, height=1in]{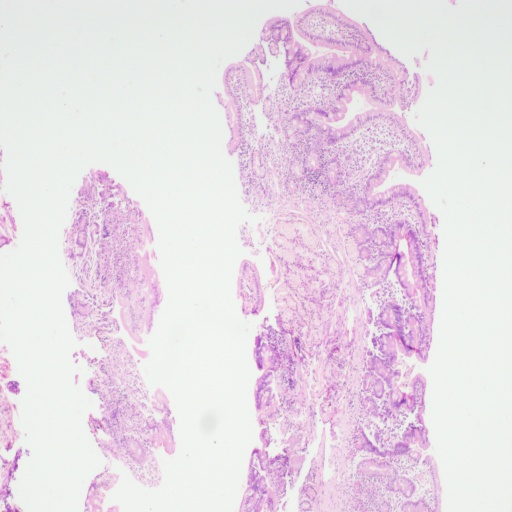}};
  \node[picture format,anchor=north]      (C1)   at (B1.south)       {\includegraphics[width=1.1in, height=1in]{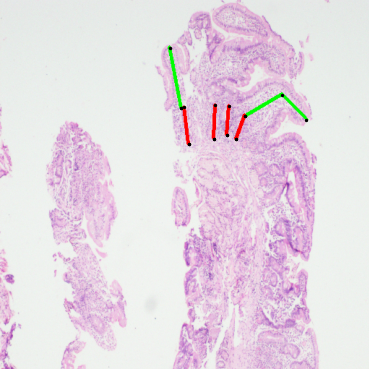}};
  \node[picture format,anchor=north]      (C2)   at (B2.south)       {\includegraphics[width=1.1in, height=1in]{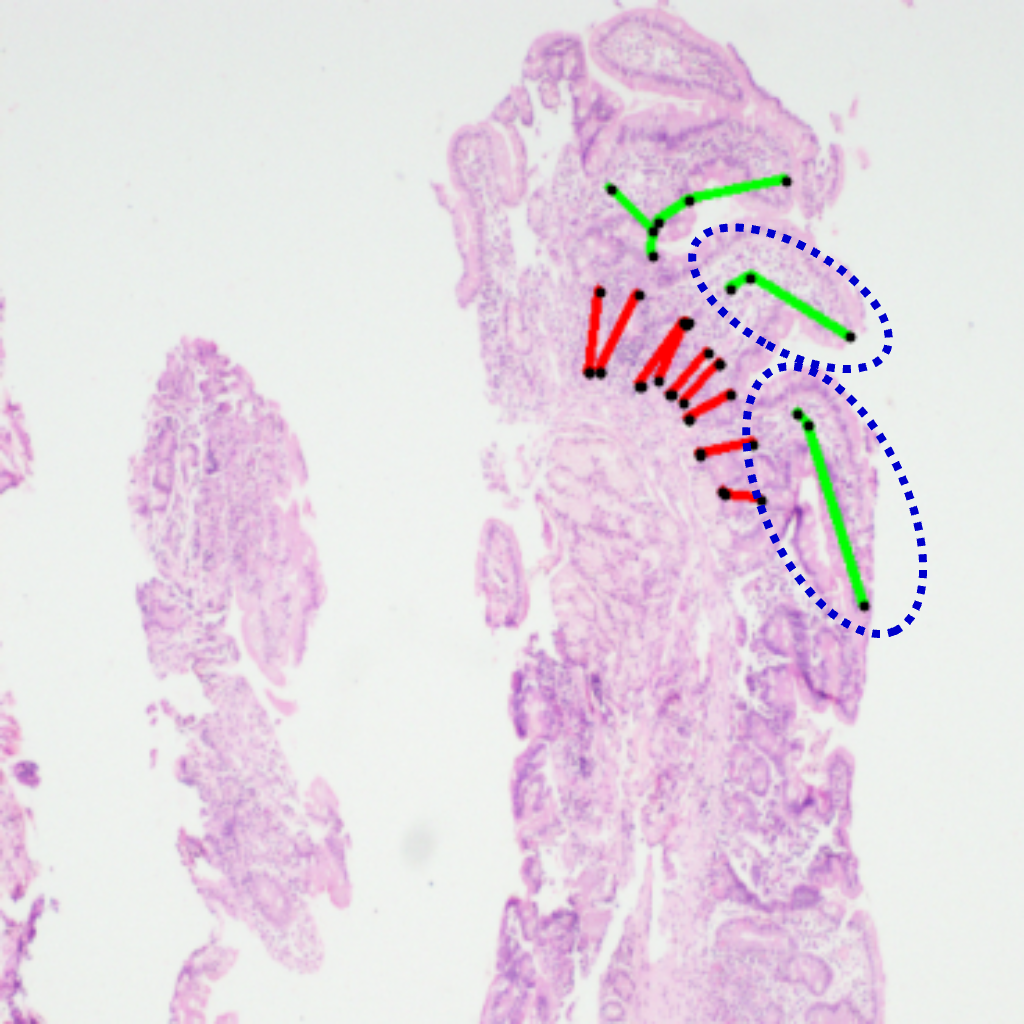}};

  \node[anchor=south] (N0) at (A0.north) {\bfseries Image};
  \node[anchor=south] (N1) at (A1.north) {\bfseries Ground Truth};
  \node[anchor=south] (N2) at (A2.north) {\bfseries MeasureNet};

\end{tikzpicture}
\caption{Three main types of errors include: 1) Missing or incorrect Villi Shoulder (VS) and Crypt Border (CB), leading to shifted crypt predictions, as shown in row 1; 2) False positive predictions for denuded villi, which should be excluded from measurements, as illustrated in row 2; and 3) Failure to capture curvature, resulting in measurement inaccuracies, as shown in row 3.}
\label{fig:Error_Analysis}
\end{figure}

We conduct an error analysis of \model{}, highlighting common errors (see Fig. \ref{fig:Error_Analysis}). Fig \ref{fig:Error_Analysis}, row 1 illustrates errors arising from mis-identification of the villi shoulder and crypt border, often due to the lack of distinct visual attributes for these regions. This results in change in ratio, and are more prominent in cases of severe CeD. Fig \ref{fig:Error_Analysis} row 2, highlights errors due to false positives villi predictions. For villi measurements pathologists consider good villi with unbroken epithelial layers (outer layer). \model{} occasionally mis-classifies denuded villi (with broken epithelial layer) as good villi. While these do not significantly affecting measurements but impacts localization performance. Fig \ref{fig:Error_Analysis} row 3 shows \model{}’s difficulty capturing sharp curvatures. Most villi can be approximated with straight polylines, leading to a bias toward linear predictions resulting in errors in villi with pronounced curvature.

\section{Conclusion}

We address the challenge of identifying celiac disease by accurately measuring villi and crypt length in duodenal biopsy images. Our proposed model, \model{}, introduces a novel polyline detection framework that combines polyline detection loss, derived from chamfer distance, with object-driven loss that includes both length and part-length loss components. To enhance crypt polyline detection, \model{} incorporates auxiliary guidance of villi shoulders and crypt borders in form of segmentation mask. Mask feature mixup prevents over-reliance on auxiliary masks, ensuring robustness and accuracy in detection.



We introduce CeDeM, a novel dataset of 750 annotated duodenal biopsy images curated for identifying celiac disease through villi and crypt measurements. \model{} demonstrates significant effectiveness on CeDeM, outperforming the closest baseline in measurement, localization, and classification metrics: achieving reduction in villi-crypt length ratio from 0.70 to 0.47, increase in mAP from 40.3 to 50.2, and improvement in classification accuracy from 70\% to 81.4\%. We will release both the code and dataset for future research.
{
    \small
    \bibliographystyle{ieeenat_fullname}
    \bibliography{main}

\begin{thebibliography}{38}
\providecommand{\natexlab}[1]{#1}
\providecommand{\url}[1]{\texttt{#1}}
\expandafter\ifx\csname urlstyle\endcsname\relax
  \providecommand{\doi}[1]{doi: #1}\else
  \providecommand{\doi}{doi: \begingroup \urlstyle{rm}\Url}\fi

\bibitem[Arora et~al.(2022)Arora, Asri, Bahuleyan, and Cheung]{arora2022exposure}
Kushal Arora, Layla~El Asri, Hareesh Bahuleyan, and Jackie Chi~Kit Cheung.
\newblock Why exposure bias matters: An imitation learning perspective of error accumulation in language generation.
\newblock \emph{arXiv preprint arXiv:2204.01171}, 2022.

\bibitem[Bao et~al.(2021)Bao, Dong, Piao, and Wei]{bao2021beit}
Hangbo Bao, Li Dong, Songhao Piao, and Furu Wei.
\newblock Beit: Bert pre-training of image transformers.
\newblock \emph{arXiv preprint arXiv:2106.08254}, 2021.

\bibitem[Cai and Hong(2018)]{cai2018quantitative}
Wen-Li Cai and Guo-Bin Hong.
\newblock Quantitative image analysis for evaluation of tumor response in clinical oncology.
\newblock \emph{Chronic diseases and translational medicine}, 4\penalty0 (01):\penalty0 18--28, 2018.

\bibitem[Carion et~al.(2020)Carion, Massa, Synnaeve, Usunier, Kirillov, and Zagoruyko]{carion2020end}
Nicolas Carion, Francisco Massa, Gabriel Synnaeve, Nicolas Usunier, Alexander Kirillov, and Sergey Zagoruyko.
\newblock End-to-end object detection with transformers.
\newblock In \emph{European conference on computer vision}, pages 213--229. Springer, 2020.

\bibitem[Carreras(2024)]{carreras2024celiac}
Joaquim Carreras.
\newblock Celiac disease deep learning image classification using convolutional neural networks.
\newblock \emph{Journal of Imaging}, 10\penalty0 (8):\penalty0 200, 2024.

\bibitem[Corazza et~al.(2007)Corazza, Villanacci, Zambelli, Milione, Luinetti, Vindigni, Chioda, Albarello, Bartolini, and Donato]{corazza2007comparison}
Gino~Roberto Corazza, Vincenzo Villanacci, Claudia Zambelli, Massimo Milione, Ombretta Luinetti, Carla Vindigni, Caterina Chioda, Luca Albarello, Daniela Bartolini, and Francesco Donato.
\newblock Comparison of the interobserver reproducibility with different histologic criteria used in celiac disease.
\newblock \emph{Clinical Gastroenterology and Hepatology}, 5\penalty0 (7):\penalty0 838--843, 2007.

\bibitem[Das et~al.(2019)Das, Gahlot, Singh, Baloda, Rawat, Verma, Khanna, Roy, George, Singh, et~al.]{das2019quantitative}
Prasenjit Das, Gaurav~PS Gahlot, Alka Singh, Vandana Baloda, Ramakant Rawat, Anil~K Verma, Gaurav Khanna, Maitrayee Roy, Archana George, Ashok Singh, et~al.
\newblock Quantitative histology-based classification system for assessment of the intestinal mucosal histological changes in patients with celiac disease.
\newblock \emph{Intestinal research}, 17\penalty0 (3):\penalty0 387--397, 2019.

\bibitem[Ensari(2010)]{ensari2010gluten}
Arzu Ensari.
\newblock Gluten-sensitive enteropathy (celiac disease): controversies in diagnosis and classification.
\newblock \emph{Archives of pathology \& laboratory medicine}, 134\penalty0 (6):\penalty0 826--836, 2010.

\bibitem[Homayounfar et~al.(2018)Homayounfar, Ma, Lakshmikanth, and Urtasun]{homayounfar2018hierarchical}
Namdar Homayounfar, Wei-Chiu Ma, Shrinidhi~Kowshika Lakshmikanth, and Raquel Urtasun.
\newblock Hierarchical recurrent attention networks for structured online maps.
\newblock In \emph{Proceedings of the IEEE Conference on Computer Vision and Pattern Recognition}, pages 3417--3426, 2018.

\bibitem[Hsiao et~al.(2022)Hsiao, Sun, Lin, Peng, Chen, Cheng, Yang, Yang, Wu, Lin, et~al.]{hsiao2022deep}
Chiu-Han Hsiao, Tzu-Lung Sun, Ping-Cherng Lin, Tsung-Yu Peng, Yu-Hsin Chen, Chieh-Yun Cheng, Feng-Jung Yang, Shao-Yu Yang, Chih-Horng Wu, Frank Yeong-Sung Lin, et~al.
\newblock A deep learning-based precision volume calculation approach for kidney and tumor segmentation on computed tomography images.
\newblock \emph{Computer methods and programs in biomedicine}, 221:\penalty0 106861, 2022.

\bibitem[Koukoutegos et~al.(2024)Koukoutegos, ’s Heeren, De~Wever, De~Keyzer, Maes, and Bosmans]{koukoutegos2024segmentation}
Konstantinos Koukoutegos, Richard ’s Heeren, Liesbeth De~Wever, Frederik De~Keyzer, Frederik Maes, and Hilde Bosmans.
\newblock Segmentation-based quantitative measurements in renal ct imaging using deep learning.
\newblock \emph{European Radiology Experimental}, 8\penalty0 (1):\penalty0 110, 2024.

\bibitem[Li et~al.(2019)Li, Li, Hu, and Yang]{li2019line}
Xiang Li, Jun Li, Xiaolin Hu, and Jian Yang.
\newblock Line-cnn: End-to-end traffic line detection with line proposal unit.
\newblock \emph{IEEE Transactions on Intelligent Transportation Systems}, 21\penalty0 (1):\penalty0 248--258, 2019.

\bibitem[Lin et~al.(2020)Lin, Goyal, Girshick, He, and Dollár]{8417976}
Tsung-Yi Lin, Priya Goyal, Ross Girshick, Kaiming He, and Piotr Dollár.
\newblock Focal loss for dense object detection.
\newblock \emph{IEEE Transactions on Pattern Analysis and Machine Intelligence}, 42\penalty0 (2):\penalty0 318--327, 2020.

\bibitem[Lin et~al.(2024)Lin, Zhou, Liu, and Zhu]{lin2024comprehensive}
Xinyu Lin, Yingjie Zhou, Yipeng Liu, and Ce Zhu.
\newblock A comprehensive review of image line segment detection and description: Taxonomies, comparisons, and challenges.
\newblock \emph{IEEE Transactions on Pattern Analysis and Machine Intelligence}, 2024.

\bibitem[Liu et~al.(2022)Liu, Hu, Lin, Yao, Xie, Wei, Ning, Cao, Zhang, Dong, et~al.]{liu2022swin}
Ze Liu, Han Hu, Yutong Lin, Zhuliang Yao, Zhenda Xie, Yixuan Wei, Jia Ning, Yue Cao, Zheng Zhang, Li Dong, et~al.
\newblock Swin transformer v2: Scaling up capacity and resolution.
\newblock In \emph{Proceedings of the IEEE/CVF conference on computer vision and pattern recognition}, pages 12009--12019, 2022.

\bibitem[Meyer et~al.(2021)Meyer, Skudlik, Pauls, and Stiller]{meyer2021yolino}
Annika Meyer, Philipp Skudlik, Jan-Hendrik Pauls, and Christoph Stiller.
\newblock Yolino: Generic single shot polyline detection in real time.
\newblock In \emph{Proceedings of the IEEE/CVF International Conference on Computer Vision}, pages 2916--2925, 2021.

\bibitem[Moon et~al.(2023)Moon, Lee, and Lee]{moon2023deep}
Ki-Ryum Moon, Byoung-Dai Lee, and Mu~Sook Lee.
\newblock A deep learning approach for fully automated measurements of lower extremity alignment in radiographic images.
\newblock \emph{Scientific Reports}, 13\penalty0 (1):\penalty0 14692, 2023.

\bibitem[Pan et~al.(2018)Pan, Shi, Luo, Wang, and Tang]{pan2018spatial}
Xingang Pan, Jianping Shi, Ping Luo, Xiaogang Wang, and Xiaoou Tang.
\newblock Spatial as deep: Spatial cnn for traffic scene understanding.
\newblock In \emph{Proceedings of the AAAI conference on artificial intelligence}, 2018.

\bibitem[Pautrat et~al.(2023)Pautrat, Barath, Larsson, Oswald, and Pollefeys]{pautrat2023deeplsd}
R{\'e}mi Pautrat, Daniel Barath, Viktor Larsson, Martin~R Oswald, and Marc Pollefeys.
\newblock Deeplsd: Line segment detection and refinement with deep image gradients.
\newblock In \emph{Proceedings of the IEEE/CVF Conference on Computer Vision and Pattern Recognition}, pages 17327--17336, 2023.

\bibitem[P{\"u}ttmann et~al.(2024)P{\"u}ttmann, Ferris, Marini, Aswolinsky, Vatrano, Fragetta, Nagtegaal, van~der Post, Ciompi, Atzori, et~al.]{puttmann2024automated}
Simon P{\"u}ttmann, Lluis~Borras Ferris, Niccol{\`o} Marini, Witali Aswolinsky, Simona Vatrano, Filippo Fragetta, Iris Nagtegaal, Chella van~der Post, Francesco Ciompi, Manfredo Atzori, et~al.
\newblock Automated classification of celiac disease in histopathological images: a multi-scale approach.
\newblock In \emph{Medical Imaging 2024: Computer-Aided Diagnosis}, pages 565--579. SPIE, 2024.

\bibitem[Ronneberger et~al.(2015)Ronneberger, Fischer, and Brox]{ronneberger2015u}
Olaf Ronneberger, Philipp Fischer, and Thomas Brox.
\newblock U-net: Convolutional networks for biomedical image segmentation.
\newblock In \emph{Medical image computing and computer-assisted intervention--MICCAI 2015: 18th international conference, Munich, Germany, October 5-9, 2015, proceedings, part III 18}, pages 234--241. Springer, 2015.

\bibitem[Sali et~al.(2019)Sali, Ehsan, Kowsari, Khan, Moskaluk, Syed, and Brown]{sali2019celiacnet}
Rasoul Sali, Lubaina Ehsan, Kamran Kowsari, Marium Khan, Christopher~A Moskaluk, Sana Syed, and Donald~E Brown.
\newblock Celiacnet: Celiac disease severity diagnosis on duodenal histopathological images using deep residual networks.
\newblock In \emph{2019 IEEE International Conference on Bioinformatics and Biomedicine (BIBM)}, pages 962--967. IEEE, 2019.

\bibitem[Sandler et~al.(2018)Sandler, Howard, Zhu, Zhmoginov, and Chen]{sandler2018mobilenetv2}
Mark Sandler, Andrew Howard, Menglong Zhu, Andrey Zhmoginov, and Liang-Chieh Chen.
\newblock Mobilenetv2: Inverted residuals and linear bottlenecks.
\newblock In \emph{Proceedings of the IEEE conference on computer vision and pattern recognition}, pages 4510--4520, 2018.

\bibitem[Sharbatdaran et~al.(2022)Sharbatdaran, Romano, Teichman, Dev, Raza, Goel, Moghadam, Blumenfeld, Chevalier, Shimonov, et~al.]{sharbatdaran2022deep}
Arman Sharbatdaran, Dominick Romano, Kurt Teichman, Hreedi Dev, Syed~I Raza, Akshay Goel, Mina~C Moghadam, Jon~D Blumenfeld, James~M Chevalier, Daniil Shimonov, et~al.
\newblock Deep learning automation of kidney, liver, and spleen segmentation for organ volume measurements in autosomal dominant polycystic kidney disease.
\newblock \emph{Tomography}, 8\penalty0 (4):\penalty0 1804--1819, 2022.

\bibitem[Tabelini et~al.(2021{\natexlab{a}})Tabelini, Berriel, Paixao, Badue, De~Souza, and Oliveira-Santos]{tabelini2021keep}
Lucas Tabelini, Rodrigo Berriel, Thiago~M Paixao, Claudine Badue, Alberto~F De~Souza, and Thiago Oliveira-Santos.
\newblock Keep your eyes on the lane: Real-time attention-guided lane detection.
\newblock In \emph{Proceedings of the IEEE/CVF conference on computer vision and pattern recognition}, pages 294--302, 2021{\natexlab{a}}.

\bibitem[Tabelini et~al.(2021{\natexlab{b}})Tabelini, Berriel, Paixao, Badue, De~Souza, and Oliveira-Santos]{tabelini2021polylanenet}
Lucas Tabelini, Rodrigo Berriel, Thiago~M Paixao, Claudine Badue, Alberto~F De~Souza, and Thiago Oliveira-Santos.
\newblock Polylanenet: Lane estimation via deep polynomial regression.
\newblock In \emph{2020 25th International Conference on Pattern Recognition (ICPR)}, pages 6150--6156. IEEE, 2021{\natexlab{b}}.

\bibitem[Tan and Le(2019)]{tan2019efficientnet}
Mingxing Tan and Quoc Le.
\newblock Efficientnet: Rethinking model scaling for convolutional neural networks.
\newblock In \emph{International conference on machine learning}, pages 6105--6114. PMLR, 2019.

\bibitem[Taniyama et~al.(2021)Taniyama, Murakami, Yoshida, Takahashi, Matsubara, Baba, and Kamei]{taniyama2021evaluating}
Yusuke Taniyama, Kentaro Murakami, Naoya Yoshida, Kozue Takahashi, Hisahiro Matsubara, Hideo Baba, and Takashi Kamei.
\newblock Evaluating the effect of neoadjuvant chemotherapy for esophageal cancer using the recist system with shorter-axis measurements: a retrospective multicenter study.
\newblock \emph{BMC cancer}, 21:\penalty0 1--10, 2021.

\bibitem[Teplyakov et~al.(2022)Teplyakov, Erlygin, and Shvets]{teplyakov2022lsdnet}
Lev Teplyakov, Leonid Erlygin, and Evgeny Shvets.
\newblock Lsdnet: Trainable modification of lsd algorithm for real-time line segment detection.
\newblock \emph{IEEE Access}, 10:\penalty0 45256--45265, 2022.

\bibitem[Tolstikhin et~al.(2021)Tolstikhin, Houlsby, Kolesnikov, Beyer, Zhai, Unterthiner, Yung, Steiner, Keysers, Uszkoreit, et~al.]{tolstikhin2021mlp}
Ilya~O Tolstikhin, Neil Houlsby, Alexander Kolesnikov, Lucas Beyer, Xiaohua Zhai, Thomas Unterthiner, Jessica Yung, Andreas Steiner, Daniel Keysers, Jakob Uszkoreit, et~al.
\newblock Mlp-mixer: An all-mlp architecture for vision.
\newblock \emph{Advances in neural information processing systems}, 34:\penalty0 24261--24272, 2021.

\bibitem[Tyagi et~al.(2023)Tyagi, Mohapatra, Das, Makharia, Mehra, AP, et~al.]{tyagi2023degpr}
Aayush~Kumar Tyagi, Chirag Mohapatra, Prasenjit Das, Govind Makharia, Lalita Mehra, Prathosh AP, et~al.
\newblock Degpr: Deep guided posterior regularization for multi-class cell detection and counting.
\newblock In \emph{Proceedings of the IEEE/CVF Conference on Computer Vision and Pattern Recognition}, pages 23913--23923, 2023.

\bibitem[Vohra(2016)]{vohra2016celiac}
Pankaj Vohra.
\newblock \emph{Celiac Disease: A Comprehensive Guide Paperback}.
\newblock National Book Trust, 2016.

\bibitem[Von~Gioi et~al.(2008)Von~Gioi, Jakubowicz, Morel, and Randall]{von2008lsd}
Rafael~Grompone Von~Gioi, Jeremie Jakubowicz, Jean-Michel Morel, and Gregory Randall.
\newblock Lsd: A fast line segment detector with a false detection control.
\newblock \emph{IEEE transactions on pattern analysis and machine intelligence}, 32\penalty0 (4):\penalty0 722--732, 2008.

\bibitem[Xie et~al.(2021)Xie, Wang, Yu, Anandkumar, Alvarez, and Luo]{xie2021segformer}
Enze Xie, Wenhai Wang, Zhiding Yu, Anima Anandkumar, Jose~M Alvarez, and Ping Luo.
\newblock Segformer: Simple and efficient design for semantic segmentation with transformers.
\newblock \emph{Advances in neural information processing systems}, 34:\penalty0 12077--12090, 2021.

\bibitem[Xu et~al.(2021)Xu, Xu, Cheung, and Tu]{xu2021line}
Yifan Xu, Weijian Xu, David Cheung, and Zhuowen Tu.
\newblock Line segment detection using transformers without edges.
\newblock In \emph{Proceedings of the IEEE/CVF Conference on Computer Vision and Pattern Recognition}, pages 4257--4266, 2021.

\bibitem[Xue et~al.(2020)Xue, Wu, Bai, Wang, Xia, Zhang, and Torr]{xue2020holistically}
Nan Xue, Tianfu Wu, Song Bai, Fudong Wang, Gui-Song Xia, Liangpei Zhang, and Philip~HS Torr.
\newblock Holistically-attracted wireframe parsing.
\newblock In \emph{Proceedings of the IEEE/CVF Conference on Computer Vision and Pattern Recognition}, pages 2788--2797, 2020.

\bibitem[Zhang(2017)]{zhang2017mixup}
Hongyi Zhang.
\newblock mixup: Beyond empirical risk minimization.
\newblock \emph{arXiv preprint arXiv:1710.09412}, 2017.

\bibitem[Zhang et~al.(2022)Zhang, Li, Liu, Zhang, Su, Zhu, Ni, and Shum]{zhang2022dino}
Hao Zhang, Feng Li, Shilong Liu, Lei Zhang, Hang Su, Jun Zhu, Lionel~M Ni, and Heung-Yeung Shum.
\newblock Dino: Detr with improved denoising anchor boxes for end-to-end object detection.
\newblock \emph{arXiv preprint arXiv:2203.03605}, 2022.

\end{thebibliography}
}

\clearpage
\setcounter{page}{1}
\maketitlesupplementary

\begin{table*}[!t]
\centering
\begin{tabular}{p{70pt} p{30pt} p{30pt} p{33pt} p{33pt}  p{30pt} p{30pt} p{40pt} p{33pt} p{30pt}}
\hline
  Model & MAE Villi $\downarrow$ & MRE Villi $\downarrow$ & MAE Crypt $\downarrow$ & MRE Crypt $\downarrow$ & MAE Ratio $\downarrow$ & MRE Ratio $\downarrow$ & Precision$\uparrow$ & Recall $\uparrow$ & mAP $\uparrow$  \\ \hline
  Middle Point  & 10.88 & 10.33 & 7.21 & 13.97  & 0.4684 & 0.2191 & 48.78 & 78.99 & 48.70 \\
  Start-End & 14.39 & 13.59 & 6.83 & 13.87 & 0.4722 & 0.2116 & 51.37 & 79.94 & 50.29\\
\hline
\end{tabular}
\caption{Comparison of measurement and detection results based on different methods for selecting the third point in polylines when the original annotation contains only two points. Middle Point refers to the midpoint of the line segment connecting the two points, while Start-End Point involves duplicating either the start or end point.}
\label{tab:point table}
\end{table*}
\begin{table*}[t]
\centering
\begin{tabular}{p{60pt} p{30pt} p{30pt} p{33pt} p{33pt}  p{30pt} p{30pt} p{40pt} p{33pt} p{30pt}}
\hline
  Model & MAE Villi $\downarrow$ & MRE Villi $\downarrow$ & MAE Crypt $\downarrow$ & MRE Crypt $\downarrow$ & MAE Ratio $\downarrow$ & MRE Ratio $\downarrow$ & Precision$\uparrow$ & Recall $\uparrow$ & mAP $\uparrow$  \\ \hline
  Earth Mover & 14.32 & 13.27 & 7.26 & 14.28 & 0.4752 & 0.212 & 47.38 & 78.86 & 48.10 \\
  Chamfer & 14.39 & 13.59 & 6.83 & 13.87 & 0.4722 & 0.2116 & 51.37 & 79.94 & 50.29\\
    \hline
\end{tabular}
\caption{Measurement and detection results for type of Distance metrics}
\label{tab:distance_Table}
\end{table*}

\begin{figure*}[t]
\centering
\begin{tikzpicture}[scale=1.0, transform shape, picture format/.style={inner sep=1pt}]

    \node[picture format]                   (A1)   at (0,0)            {\includegraphics[width=1.4in, height=1.6in]{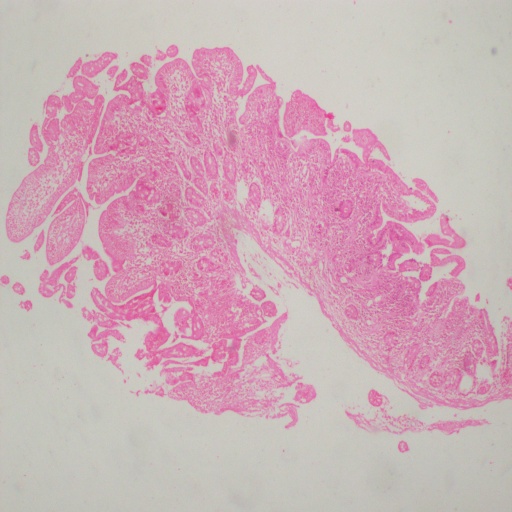}};
    \node[picture format,anchor=north]      (B1) at (A1.south)         {\includegraphics[width=1.4in, height=1.6in]{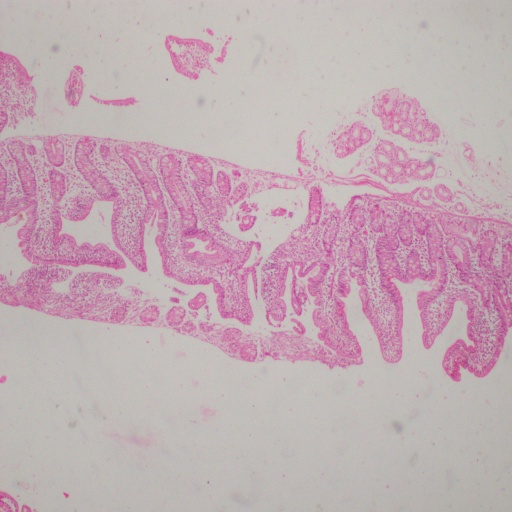}};
    \node[picture format,anchor=north]      (C1) at (B1.south)         {\includegraphics[width=1.4in, height=1.6in]{Images/Figure6_Sup/5_Image_Detection.jpg}};
    \node[picture format,anchor=north]      (D1) at (C1.south)         {\includegraphics[width=1.4in, height=1.6in]{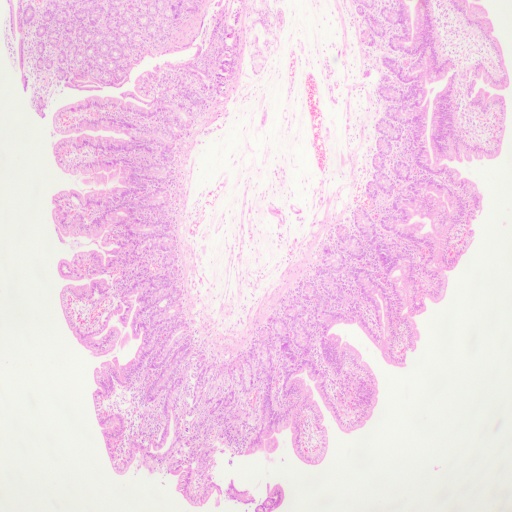}};

    \node[picture format,anchor=north west] (A2) at (A1.north east)    {\includegraphics[width=1.4in, height=1.6in]{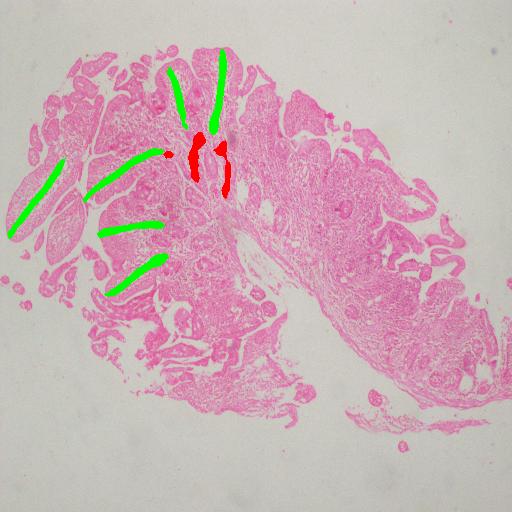}};
    \node[picture format,anchor=north]      (B2) at (A2.south)         {\includegraphics[width=1.4in, height=1.6in]{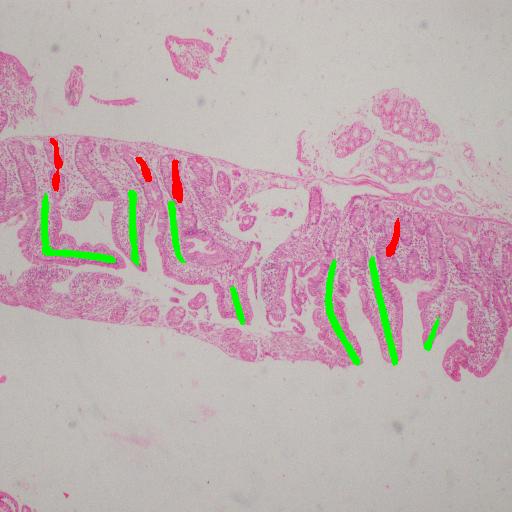}};
    \node[picture format,anchor=north]      (C2) at (B2.south)         {\includegraphics[width=1.4in, height=1.6in]{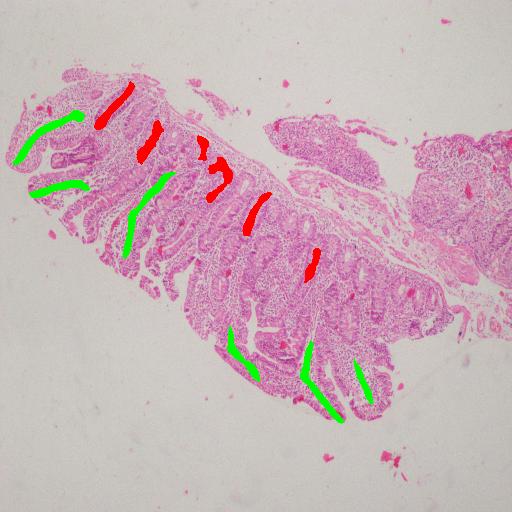}};
    \node[picture format,anchor=north]      (D2) at (C2.south)         {\includegraphics[width=1.4in, height=1.6in]{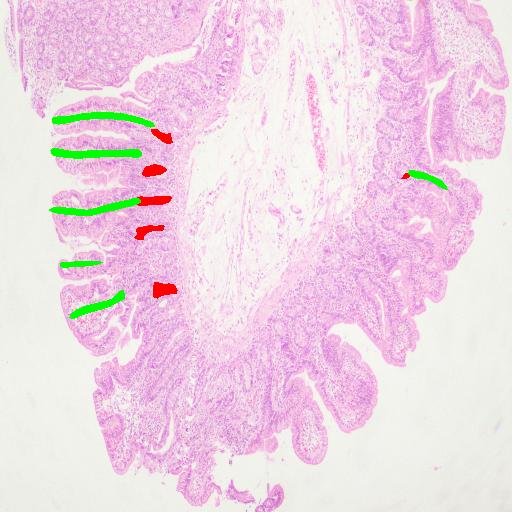}};

    \node[picture format,anchor=north west] (A3) at (A2.north east)    {\includegraphics[width=1.4in, height=1.6in]{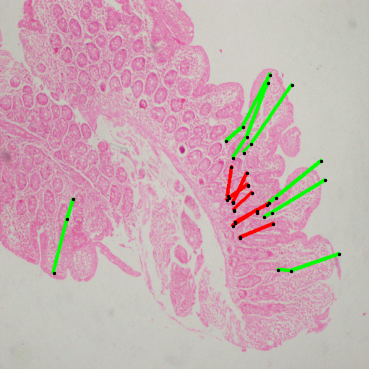}};
    \node[picture format,anchor=north]      (B3) at (A3.south)         {\includegraphics[width=1.4in, height=1.6in]{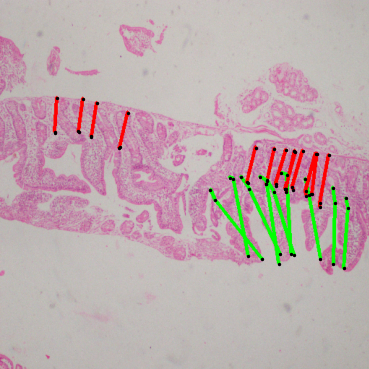}};
    \node[picture format,anchor=north]      (C3) at (B3.south)         {\includegraphics[width=1.4in, height=1.6in]{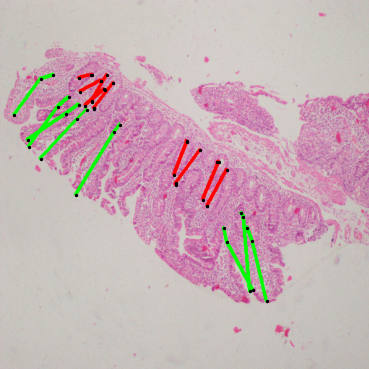}};
    \node[picture format,anchor=north]      (D3) at (C3.south)         {\includegraphics[width=1.4in, height=1.6in]{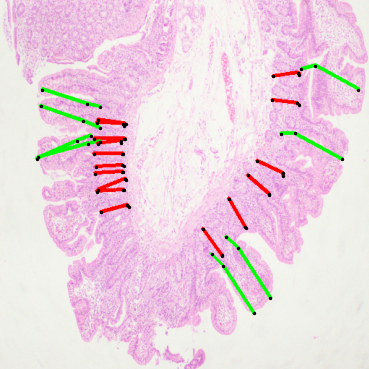}};

    \node[picture format,anchor=north west] (A4) at (A3.north east)    {\includegraphics[width=1.4in, height=1.6in]{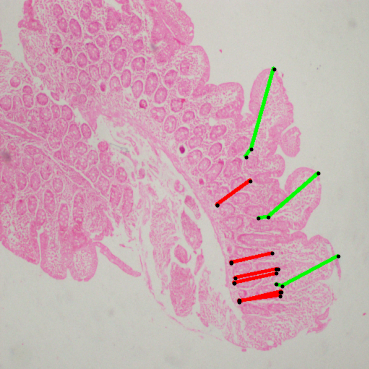}};
    \node[picture format,anchor=north]      (B4) at (A4.south)         {\includegraphics[width=1.4in, height=1.6in]{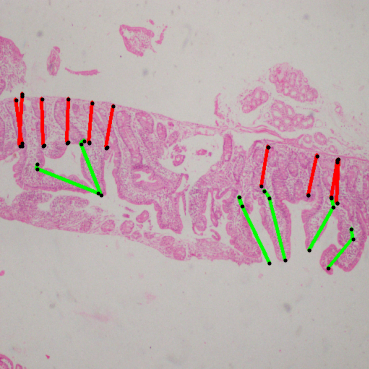}};
    \node[picture format,anchor=north]      (C4) at (B4.south)         {\includegraphics[width=1.4in, height=1.6in]{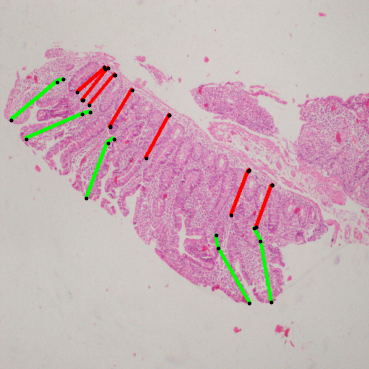}};
    \node[picture format,anchor=north]      (D4) at (C4.south)         {\includegraphics[width=1.4in, height=1.6in]{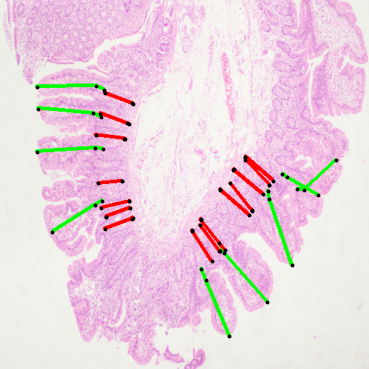}};

    \node[picture format,anchor=north west] (A5) at (A4.north east)    {\includegraphics[width=1.4in, height=1.6in]{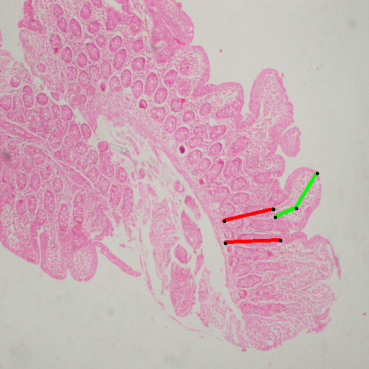}};
    \node[picture format,anchor=north]      (B5) at (A5.south)         {\includegraphics[width=1.4in, height=1.6in]{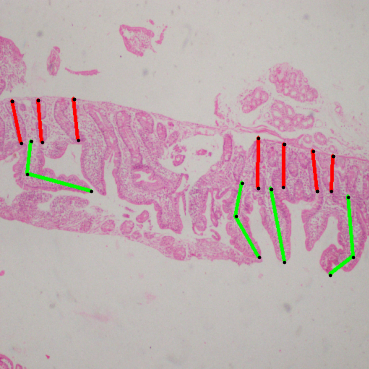}};
    \node[picture format,anchor=north]      (C5) at (B5.south)         {\includegraphics[width=1.4in, height=1.6in]{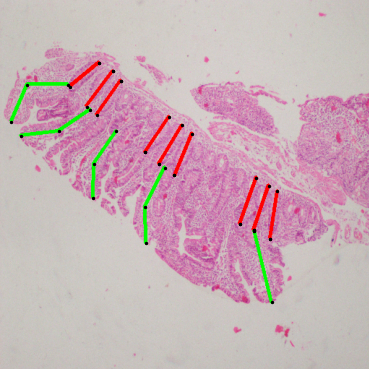}};
    \node[picture format,anchor=north]      (D5) at (C5.south)         {\includegraphics[width=1.4in, height=1.6in]{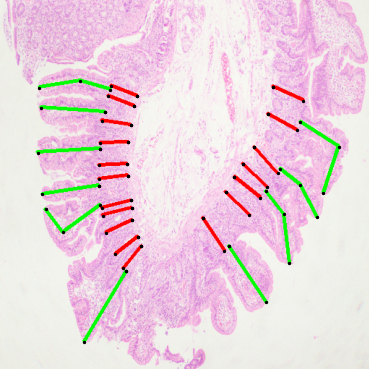}};

    \node[anchor=south] (N1) at (A1.north) {\bfseries Image};
    \node[anchor=south] (N2) at (A2.north) {\bfseries Segmentation};
    \node[anchor=south] (N3) at (A3.north) {\bfseries LETR};
    \node[anchor=south] (N4) at (A4.north) {\bfseries MeasureNet};
    \node[anchor=south] (N5) at (A5.north) {\bfseries Ground Truth};

\end{tikzpicture}
\caption{Qualitative performance comparison of various models for villi and crypt detection and localization. Columns show the original image, SegFormer predictions, LETR predictions, MeasureNet predictions, and ground truth, respectively. Green and red represent villi and crypt annotations.}
\label{fig:Detection_Visualization_Sup}
\end{figure*}

\section{Ablation Studies}
\subsection{Effect of Point Selection}
We used three points to represent a polyline, replicating the first or last point when the original annotation contained only two points. Alternatively, the middle point of the line segment can be used for polylines with two points. In Table \ref{tab:point table}, we compare both approaches. Our results show that using the middle point improves the curvature representation for villi, enabling a more accurate depiction of their natural shape. However, for crypts, adding a middle point can introduce a slight curvature or bend, leading to errors in measurement.



\subsection{Effect of Distance Metric}

We compare the performance of MeasureNet using Chamfer Distance and Earth Mover’s Distance (EMD) in Table \ref{tab:distance_Table}. Earth Mover’s Distance (EMD), also known as Wasserstein Distance, measures the minimum cost required to transform one probability distribution into another, accounting for both point distances and weights. Our experiments show that MeasureNet with Chamfer Distance achieves the best performance.



\section{Implementation Details}

Yolino, originally designed for single-class lane predictions, was adapted for our task by training two separate models: one for villi and another for crypts. The polyline annotations were reformatted to align with lane detection requirements, and each image was divided into a $32 \times 32$ grid to ensure that the start and end coordinates of the polyline fell within the grid. Yolino was trained for 80 epochs using the Adam optimizer with a learning rate of $1 \times 10^{-4}$.
We trained BEiT using the \textit{beit-base-finetuned-ade-640-640} backbone for 200 epochs with a learning rate of $1 \times 10^{-6}$ and a batch size of 4. For LETR, we employed DINO-DETR as the base architecture with a Swin Transformer backbone. LETR was trained for 150 epochs with a batch size of 2, a learning rate of $1 \times 10^{-4}$, a query dimension of 4, and number of queries being 900.

\end{document}